\documentclass[10pt,journal,letterpaper,final]{IEEEtran}
\usepackage{times}
\usepackage{epsfig}
\usepackage{float}
\usepackage{graphicx}
\usepackage{amsmath}
\usepackage{amssymb}
\usepackage{amsthm}
\usepackage[linesnumbered,ruled,vlined]{algorithm2e}
\usepackage{subfigure}
\usepackage{colortbl,booktabs}
\usepackage{threeparttable}
\usepackage{subeqnarray}
\usepackage{cases}
\usepackage{multirow}
\usepackage{array}
\usepackage[table]{xcolor}

\newcolumntype{I}{!{\vrule width 2pt}}
\newlength\savedwidth
\newcommand\whline{\noalign{\global\savedwidth\arrayrulewidth
                            \global\arrayrulewidth 2pt}%
                   \hline
                   \noalign{\global\arrayrulewidth\savedwidth}}
\newlength\savewidth

\DeclareMathOperator*{\argmin}{argmin}

\newtheorem{theorem}{Theorem}
\newtheorem{lemma}{Lemma}

\theoremstyle{definition}
\newtheorem*{pf}{Proof}

\ifCLASSOPTIONcompsoc
\else
\fi
\ifCLASSINFOpdf
\else
\fi
\begin{document}
%
\title{Weighted Schatten $p$-Norm Minimization for Image Denoising and Background Subtraction}
%
%
%
%

\author{Yuan~Xie,~\IEEEmembership{Member,~IEEE,}
        Shuhang~Gu,
        Yan Liu,
        Wangmeng~Zuo,
        Wensheng~Zhang,
        and Lei~Zhang,~\IEEEmembership{Senior~Member,~IEEE}
\IEEEcompsocitemizethanks{
\IEEEcompsocthanksitem Y. Xie is with the Department of Computing, The Hong Kong Polytechnic University, Hong Kong, China, and also with the Research Center of Precision Sensing and Control, Institute of Automation, Chinese Academy of Sciences, Beijing, 100190, China; E-mail: yuan.xie@ia.ac.cn.
\IEEEcompsocthanksitem S. Gu, Y. Liu and L. Zhang are with the Department of Computing, The Hong Kong Polytechnic University, Hong Kong, China; E-mail: shuhanggu@gmail.com, csyliu@comp.polyu.edu.hk, cslzhang@comp.polyu.edu.hk.
\IEEEcompsocthanksitem W. Zuo is with the School of Computer Science and Technology, Harbin Institute of Technology, Harbin, 150001, China; E-mail: cswmzuo@gmail.com.
\IEEEcompsocthanksitem W. Zhang is with the Research Center of Precision Sensing and Control, Institute of Automation, Chinese Academy of Sciences, Beijing, 100190, China; E-mail: wensheng.zhang@ia.ac.cn}
\thanks{}}

%
%

\markboth{SUBMIT TO IEEE TRANSACTIONS ON IMAGE PROCESSING}%
{Shell \MakeLowercase{\textit{et al.}}: Bare Demo of IEEEtran.cls for Computer Society Journals}
%


\IEEEcompsoctitleabstractindextext{%
\begin{abstract}
Low rank matrix approximation (LRMA), which aims to recover the underlying low rank matrix from its degraded observation, has a wide range of applications in computer vision. The latest LRMA methods resort to using the nuclear norm minimization (NNM) as a convex relaxation of the nonconvex rank minimization. However, NNM tends to over-shrink the rank components and treats the different rank components equally, limiting its flexibility in practical applications. We propose a more flexible model, namely the Weighted Schatten $p$-Norm Minimization (WSNM), to generalize the NNM to the Schatten $p$-norm minimization with weights assigned to different singular values. The proposed WSNM not only gives better approximation to the original low-rank assumption, but also considers the importance of different rank components. We analyze the solution of WSNM and prove that, under certain weights permutation, WSNM can be equivalently transformed into independent non-convex $l_p$-norm subproblems, whose global optimum can be efficiently solved by generalized iterated shrinkage algorithm. We apply WSNM to typical low-level vision problems, e.g., image denoising and background subtraction. Extensive experimental results show, both qualitatively and quantitatively, that the proposed WSNM can more effectively remove noise, and model complex and dynamic scenes compared with state-of-the-art methods.
\end{abstract}

}

\maketitle

\IEEEdisplaynotcompsoctitleabstractindextext

%
\IEEEpeerreviewmaketitle

\section{Introduction}
There is a rapidly growing interest in the recovery of an unknown low rank matrix from its degraded observation, namely low rank matrix approximation (LRMA). For example, the video clip captured by a static camera satisfies the ``low rank + sparse'' structure so that the background modeling can be conducted via LRMA \cite{RPCA,LRRP}. Also, the occluded or corrupted faces can be recovered by exploiting the low rank nature of matrix constructed by human facial images \cite{Torre,LRR,LRMAL1}. The success of recent image restoration methods \cite{BM3D,LSSC,SAIST} partly stem from nonlocal self-similarities in natural images, which can be also considered as a low rank priori. Thanks to the current convex/non-convex optimization techniques, a large amount of modified models and improved algorithms have been proposed for LRMA \cite{RPCA,LRMA1,LRMA2,LRMA3,schatten-p-1,irnn,irls-p}.

Speaking, LRMA can be achieved by factorization based models \cite{LRMF1,LRMF2,LRMF3,LRMF4} and regularization based models. We focus on the latter category in this work. One of the most representative low rank regularizers is the nuclear norm, which is defined as the sum of the singular values of a given matrix $\mathbf{X}\in \mathbb{R}^{m\times n}$, {\it i.e.}, $\|\mathbf{X}\|_\ast = \sum_{i}|\sigma_{i}(\mathbf{X})|_1$. According to \cite{tightest-convex}, nuclear norm is the tightest convex relaxation of the original rank minimization problem. Given a matrix $\mathbf{Y}$, the aim of nuclear norm minimization (NNM) is to find a low rank matrix $\mathbf{X}$ which satisfies the following objective function:
\begin{equation}\label{nnm}
    \hat{\mathbf{X}} = \arg \min_{\mathbf{X}} \| \mathbf{X} -\mathbf{Y} \|_{F}^{2}
                         + \lambda\| \mathbf{X} \|_{\ast},
\end{equation}
where $\lambda$ is a trade-off parameter between the loss function and the low rank regularization induced by the nuclear norm. Candes {\it et al.} \cite{NNM-therom} showed that the low rank matrix can be perfectly recovered by NNM under certain conditions, and Cai {\it et al.} \cite{svt} proposed a soft-thresholding operation to solve NNM efficiently. Due to the theoretical guarantees and the simple optimization scheme, NNM has been attracting great research interests in recent years.

Despite the convexity of the NNM model, it has been indicated in \cite{schatten-p-1,irls-p,schatten-p-2,non-convex-relax1} that the recovery performance of such a convex relaxation will degrade in the presence of measurement noise, and the solution can seriously deviate from the original solution of rank minimization problem. More specifically, as shown in the experimental section, the NNM based model will shrink too much the low rank components of the data. Therefore, it has been proposed in \cite{schatten-p-1,irls-p,schatten-p-2,gsvd} to enforce low rank regularization by using the schatten $p$-norm, which is defined as the $l_p$ norm of the singular values $(\sum_{i} \sigma_{i}^{p})^{1/p}$ with $0<p\leq 1$. Theoretically, schatten $p$-norm will guarantee a more accurate recovery of the signal while requiring only a {\it weaker restricted isometry property} than traditional trace norm \cite{schatten-p-2}. The empirical results also show that the schatten $p$-norm based model outperforms the standard NNM. However, most of the schatten $p$-norm based models treat all singular values equally, and they are not flexible enough to deal with many real problems where different rank components have different importances.

Another way to improve the performance of low rank approximation is to treat each rank component differently other than treat the singular values equally as in NNM. The truncated nuclear norm regularization (TNNR) \cite{tnnr} and the partial sum minimization (PSM) \cite{psm} resort to only minimizing the smallest $N - r$ singular values while keeping the largest $r$ ones unchanged, where $N$ is the number of singular values and $r$ is the rank of the matrix. Nevertheless, the rank $r$ is hard to estimate, and could vary with the content of the data matrix. To incorporate the prior knowledge of different singular values more reasonably, recently Gu {\it et al.} \cite{wnnm} proposed the weighted nuclear norm, which is defined as $\|\mathbf{X}\|_{\mathbf{w}, \ast} = \sum_{i}|w_i \sigma_{i}(\mathbf{X})|_1$, where $\mathbf{w} = [w_1,\ldots,w_n]$ and $w_i \geq 0$ is a non-negative weight assigned to $\sigma_i(\mathbf{X})$. The weighted nuclear norm is flexible to deal with many real problems, {\it e.g.}, image restoration, in which larger singular values need to be penalized less than smaller ones so that the major data components can be preserved. Compared with traditional NNM, the so-called weighted nuclear norm minimization (WNNM) scheme assigns different weights to different singular values such that the values of soft thresholds become more reasonable.

Inspired by the schatten $p$-norm minimization \cite{schatten-p-1} and WNNM \cite{wnnm}, in this paper we propose a new low rank regularizer namely \textbf{W}eighted \textbf{S}chatten $p$-\textbf{N}orm \textbf{M}inimization (WSNM) for LRMA. WSNM introduces flexibility in dealing with different rank components, and gives better approximation to the original LRMA problem. As can be seen later, WSNM generalizes WNNM to a unified low rank based framework, while WNNM is a special case of the proposed WSNM.

Nonetheless, introducing weights for schatten $p$-norm makes the problem much more difficult than the one contains only weighted schema or schatten $p$-norm, since the weights order plays a critical role in optimization. Without using weighted schema, the problem can be directly decomposed into independent $l_p$-norm subproblems \cite{gsvd}. But such a solution does not fit our case due to the auxiliary weight variables. Choosing $p=1$, the optimal solution can be achieved by the widely used soft-thresholding operator when the weights satisfy a non-descending order \cite{wnnm}. To solve our problem, we first present the general solution of WSNM, and then show that under certain weights permutation, WSNM can be equivalently transformed into independent non-convex $l_p$-norm subproblems, whose global optimum can be efficiently solved by the recently proposed generalized soft-thresholding algorithm (GST) \cite{gisa}. Meanwhile, rigorous mathematical proof of the equivalence is proved rigorously by analyzing the property of GST. In summary, we highlight the main contributions of this paper as follows:

\begin{itemize} \setlength{\itemsep}{2pt}
\item We propose a new model of LRMA, namely \textbf{W}eighted \textbf{S}chatten $p$-\textbf{N}orm \textbf{M}inimization (WSNM), and present its general solution.
\item We present an efficient optimization algorithm to solve WSNM with non-descending weights, which is supported by a rigorous proof.
\item We apply the proposed WSNM to image denoising and background subtraction, and achieve state-of-the-art performance, which demonstrates the great potentials of WSNM in low level vision applications.
\end{itemize}

The remainder of this paper is organized as follows. In Section \ref{WSNM}, we describe our proposed WSNM in details and analyze the optimization scheme to solve it. In Section \ref{WSNM-Vision}, the WSNM and its derived model WSNM-RPCA are applied to image denosing and background substraction, respectively. The experimental results are demonstrated in Section \ref{experiment}, and we conclude the proposed methods in Section \ref{discussion-and-conclusion}.

\section{Weighted Schatten $p$-Norm Minimization}\label{WSNM}

\begin{figure*}[htbp]
\setlength{\abovecaptionskip}{0pt}  
\setlength{\belowcaptionskip}{0pt} 
\renewcommand{\figurename}{Figure}
\centering
\includegraphics[width=0.9\textwidth]{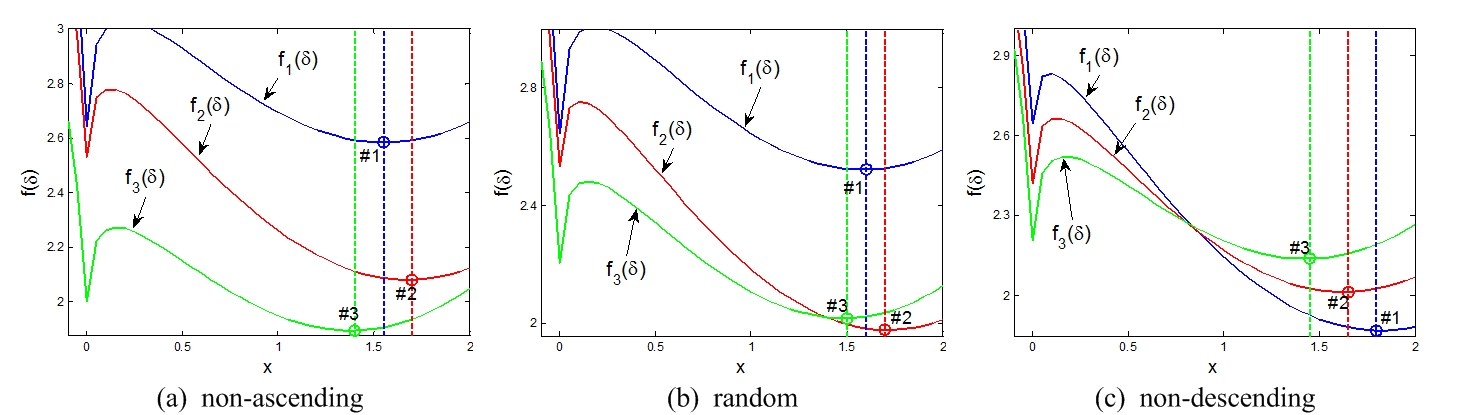}
\caption{Plots of the function $f_i(\delta)$ with different orders of $w_i$. In each subfigures, points $\#1, \#2$ and $\#3$ denote the global optimums of $f_1(\delta), f_2(\delta)$ and $f_3(\delta)$, respectively. (a) $\sigma_1 = 2.3, \sigma_2 = 2.25, \sigma_3 = 2.0$, and $w_1 = 1.85, w_2 = 1.48, w_3 = 1.45$, $\#3\leq\#1\leq\#2$ (compare the horizontal coordinate). (b) $\sigma_1 = 2.3, \sigma_2 = 2.25, \sigma_3 = 2.1$, and $w_1 = 1.8, w_2 = 1.4, w_3 = 1.5$, $\#3\leq\#1\leq\#2$. (c) $\sigma_1 = 2.3, \sigma_2 = 2.2, \sigma_3 = 2.1$, and $w_1 = 1.3, w_2 = 1.45, w_3 = 1.6$, $\#3\leq\#2\leq\#1$.}
\label{fig:w-and-y}
\end{figure*}

\subsection{Problem Formulation}

The proposed weighted schatten $p$-norm of matrix $\mathbf{X}\in \mathbb{R}^{m\times n}$ is defined as
\begin{equation}\label{}
    \|\mathbf{X}\|_{\mathbf{w},S_p} = \Bigg( \sum\nolimits_{i=1}^{\min\{ n,m \}} w_i \sigma_i^{p} \Bigg)^{\frac{1}{p}},
\end{equation}
where $\mathbf{w} = [w_1, \ldots, w_{\min\{ n,m \}}]$ is a non-negative vector, and $\sigma_i$ is the $i$-th singular value of $\mathbf{X}$. Then the weighted schatten $p$-norm of a matrix $\mathbf{X}$ to power $p$ is
\begin{equation}\label{}
    \|\mathbf{X}\|_{\mathbf{w},S_p}^{p} = \sum\nolimits_{i=1}^{\min\{ n,m \}} w_i \sigma_i^{p}
                               = tr(\mathbf{W}\mathbf{\Delta}^{p}),
\end{equation}
where both $\mathbf{W}$ and $\mathbf{\Delta}$ are diagonal matrices whose diagonal entries are composed of $w_i$ and $\sigma_i$, respectively.

Given a matrix $\mathbf{Y}$, our proposed LRMA model aims to find a matrix $\mathbf{X}$, which is as close to $\mathbf{Y}$ as possible under the F-norm data fidelity and the weighted schatten $p$-norm regularization:
\begin{equation}\label{patch-model1}
    \hat{\mathbf{X}} = \arg \min_{\mathbf{X}} \| \mathbf{X} -\mathbf{Y} \|_{F}^{2}
                         + \| \mathbf{X} \|_{w, S_p}^{p}.
\end{equation}
The solution of the above weighted schatten $p$-norm minimization (WSNM) problem is discussed in detail in the next subsection. Note that WNNM \cite{wnnm} is a special case of WSNM when power $p$ is set to $1$.

\subsection{General Solution of WSNM}

Having discussed in \cite{wnnm}, the convexity property of the optimization problem can not be preserved because of the added weights in NNM. Furthermore, the nonconvex relaxation brought by the schatten $p$-norm makes the above problem much more difficult to optimize. We first give the following theorem and lemma before analyzing the optimization of WSNM:
\begin{theorem}\label{von-neumann}
  (Von-Neumann \cite{von-neumann}) For any $m\times n$ matrices $\mathbf{A}$ and $\mathbf{B}$, $\sigma(\mathbf{A}) = [\sigma_{1}(\mathbf{A}), \ldots, \sigma_{r}(\mathbf{A})]^{T}$ and $\sigma(\mathbf{B}) = [\sigma_{1}(\mathbf{B}), \ldots, \sigma_{r}(\mathbf{B})]^{T}$, where $r = \min(m, n)$, are the singular values of $\mathbf{A}$ and $\mathbf{B}$ respectively, then $tr(\mathbf{A}^{T}\mathbf{B}) \leq tr(\sigma(\mathbf{A})^{T}\sigma(\mathbf{B}))$. The case of equality occurs if and only if it is possible to find unitaries $\mathbf{U}$ and $\mathbf{V}$ that simultaneously singular value decompose $\mathbf{A}$ and $\mathbf{B}$ in the sense that
  \begin{equation}\label{equality-hold}
    \mathbf{A} = \mathbf{U}\mathbf{\Sigma}_{A}\mathbf{V}^{T}, \text{and  } \mathbf{B} = \mathbf{U}\mathbf{\Sigma}_{B}\mathbf{V}^{T},
  \end{equation}
  where $\mathbf{\Sigma}_{A}$ and $\mathbf{\Sigma}_{B}$ denote ordered eigenvalue matrices with the singular values $\sigma(\mathbf{A})$ and $\sigma(\mathbf{B})$ along the diagonal with the same order, respectively.
\end{theorem}

\begin{lemma}\label{solution-lemma}
  Let the SVD of $\mathbf{Y}$ be $\mathbf{Y} = \mathbf{U}\mathbf{\Sigma} \mathbf{V}^{T}$ with $\mathbf{\Sigma} = diag(\sigma_1, \ldots, \sigma_r)$. Suppose that all the singular values are in \textbf{non-ascending} order, then the optimal solution of (\ref{patch-model1}) will be $\mathbf{X} = \mathbf{U}\mathbf{\Delta} \mathbf{V}^{T}$ with $\mathbf{\Delta} = diag(\delta_1, \ldots, \delta_r)$, where $\delta_i$ is given by solving the problem below:

  \begin{subequations}\label{equivalence-problem}
    \begin{numcases}{}
        \min_{\delta_1, \ldots, \delta_r} \sum_{i=1}^{r} \Bigg[(\delta_i - \sigma_i)^{2} + w_i \delta_i^{p}\Bigg], i=1,\ldots, r \label{equivalence-problem1} \\
        \text{s.t. } \delta_i \geq 0, \text{ and } \delta_i \geq \delta_j, \text{ for } i\leq j. \label{equivalence-problem2}
    \end{numcases}
  \end{subequations}
\end{lemma}

\begin{pf}
  The proof can be found in the appendix.
\end{pf}

Even transform the problem (\ref{patch-model1}) to the reduced (\ref{equivalence-problem}), solving the problem (\ref{equivalence-problem}) is still non-trivial because of the non-convexity and non-smoothness of the objective function with additional order constraint ({\it i.e.}, $\delta_i \geq \delta_j$, $i\leq j$). Intuitively, if the order constraint in (\ref{equivalence-problem2}) can be discarded, the problem (\ref{equivalence-problem1}) can be consequently decoupled into $r$ independent subproblems:
\begin{equation}\label{equivalence-problem-decoupled}
  \min_{\delta_i \geq 0} f_i(\delta) = (\delta_i - \sigma_i)^{2} + w_i \delta_i^{p}, i=1, \dots, r,
\end{equation}
then the challenges can be much reduced. The schatten $p$-norm based minimization has been discussed in several recent works such as \cite{nikolova,gisa}. Before analyzing the solution of problem (\ref{equivalence-problem}), here we firstly introduce an efficient solution of its partial problem (\ref{equivalence-problem-decoupled}). Without loss of generality, the non-negative constraint $\delta \geq 0$ can be dropped since singular value $\sigma_i \geq 0$ \cite{gisa}. Each subproblem can be effectively solved by the generalized soft-thresholding (GST) algorithm proposed in \cite{gisa} (Algorithm \ref{GST}). Given $p$ and $w_i$, there exists a specific threshold
\begin{equation}\label{generalized-threshold}
  \tau^{GST}_{p}(w_i) = (2w_i(1-p))^{\frac{1}{2-p}} + w_i p(2w_i (1-p))^{\frac{p-1}{2-p}}.
\end{equation}
If $\sigma_i< \tau^{GST}_{p}(w_i)$, $\delta_i=0$ is the global minimum; otherwise, the optimum will be obtained at non-zero point. According to \cite{gisa}, for any $\sigma_i \in (\tau^{GST}_{p}(w_i), +\infty)$, $f_i(\delta)$ has one unique minimum $S^{GST}_p(\sigma_i;w_i)$, which can be obtained by solving the following equation:
\begin{equation}\label{generalized-threshold}
  S^{GST}_p(\sigma_i;w_i) - \sigma_i + w_i p \bigg(S^{GST}_p(\sigma_i;w_i)\bigg)^{p-1} = 0.
\end{equation}
For more details about the GST algorithm, please refer to \cite{gisa}. Here, we can consider the global minimum $S^{GST}_p(\sigma;w)$ as an implicit function w.r.t variables $\sigma$ and $w$. By analyzing the property of $S^{GST}_p(\sigma;w)$, we can achieve efficient solution for WSNM under certain weights permutation, which will be discussed in the next subsection.

\SetAlFnt{\footnotesize}{ 
\begin{algorithm}[]\label{GST}
\SetAlgoLined
\caption{Generalized Soft-Thresholding (GST) \cite{gisa}}
\KwIn{$\sigma$, $w$, $p$, $J$}
\BlankLine
$\tau^{GST}_{p}(w) = (2w(1-p))^{\frac{1}{2-p}} + w p(2w (1-p))^{\frac{p-1}{2-p}}$\;
\eIf{$|\sigma| \leq \tau^{GST}_{p}(w)$}
{
    $S^{GST}_p(\sigma;w) = 0$\;
}
{
    $k=0$, $\delta^{(k)} = |\sigma|$\;
    \For{$k=0,1,\ldots,J$}
    {
        $\delta^{(k+1)} = |\sigma| - wp\big( \delta^{(k)} \big)^{p-1}$\;
        $k = k+1$\;
    }
    $S^{GST}_p(\sigma;w) = \text{sgn}(\sigma)\delta^{(k)}$\;
}

\textbf{Return} $S^{GST}_p(y;w)$\;
\end{algorithm}}

\subsection{Efficient Solution with Non-descending Weights}\label{solver-weight}
We now return to the optimization problem (\ref{equivalence-problem}). Unfortunately, the solutions of the decoupled subproblems (\ref{equivalence-problem-decoupled}) may not satisfy the order constraint (\ref{equivalence-problem2}) due to the influence of different weight conditions, which is illustrated in Fig. {\ref{fig:w-and-y}}. Suppose that $\sigma_i$ ($i=1,\ldots, r$) are in non-ascending order, from Fig. {\ref{fig:w-and-y}} (a) and Fig. {\ref{fig:w-and-y}} (b), the horizontal coordinates of $\delta_1$ ($\#1$), $\delta_2$ ($\#2$) and $\delta_3$ ($\#3$) reveal that the order constraint in (\ref{equivalence-problem2}) will not hold if $w_i$ ($i=1,\ldots, r$) are non-ascending or in arbitrary order. However, Fig. {\ref{fig:w-and-y}} (c) ($\#3\leq\#2\leq\#1$) gives us a hypothesis that constraint (\ref{equivalence-problem2}) can be satisfied in the case of non-descending order of weights $w_i$. To validate this hypothesis, we introduce the following lemma:

\begin{lemma}\label{my-lemma}
  Considering the unique minimum $S^{GST}_p(\sigma;w)$ as an implicit function {\it w.r.t.} $\sigma$ and $w$, we have the following inequality when $\sigma$ is fixed:
  \begin{equation}\label{lemma-object}
    S^{GST}_p(\sigma;w_i) \geq S^{GST}_p(\sigma;w_j), \text{ for } w_i \leq w_j, i \leq j
  \end{equation}
\end{lemma}
\begin{pf}
  The proof can be found in the appendix.
\end{pf}

Given the above lemma, we are now ready to prove the following Theorem \ref{my-theorem}:

\begin{theorem}\label{my-theorem}
  If the weights satisfy $0\leq w_1 \leq w_2 \leq \ldots \leq w_r$, the optimal solutions of all the independent subproblems in (\ref{equivalence-problem-decoupled}) also satisfy the order constraint $\delta_1 \geq \delta_2 \geq \ldots \geq \delta_r$.
\end{theorem}
\begin{pf}
  The proof can be also found in the appendix.
\end{pf}

According to Theorem \ref{my-theorem}, when the weights are in non-descending order, solving all the decoupled subproblems (\ref{equivalence-problem-decoupled}) is equivalent to solve the coupled problem (\ref{equivalence-problem}). So far, the original problem (\ref{patch-model1}) has been converted to (\ref{equivalence-problem-decoupled}) which can be solved more easily. The proposed algorithm of WSNM for LRMA is summarized in Algorithm \ref{propose-wsnm}. Generally, the non-descending order of $w_i$ is meaningful for most practical applications in computer vision, because the components with larger singular values need to be penalized less than smaller ones, and hence the preservation of major data components can be guaranteed.

\SetAlFnt{\footnotesize}{ 
\begin{algorithm}[]\label{propose-wsnm}
\SetAlgoLined
\caption{WSNM via GST}
\KwIn{$\mathbf{Y}$, $\{w_i\}_{i=1}^{r}$ in non-descending order, $p$}
\KwOut{Matrix $\hat{\mathbf{X}}$\\}
\BlankLine
$\mathbf{Y} = \mathbf{U}\mathbf{\Sigma}\mathbf{V}^{T}$, $\mathbf{\Sigma} = \text{diag}(\sigma_1, \ldots, \sigma_r)$\;
\For(\tcc*[f]{can calculate in parallel}){$i=1$ \KwTo $r$}
{
    $\delta_i$ = \text{GST}($\sigma_i$, $w_i$, $p$)\;
}
$\mathbf{\Delta} = \text{diag}(\delta_1, \ldots, \delta_r)$\;
\textbf{Return} $\hat{\mathbf{X}} = \mathbf{U} \mathbf{\Delta} \mathbf{V}^{T}$\;
\end{algorithm}}

\section{Applying WSNM to Image Denoising and Background Substraction}\label{WSNM-Vision}
To validate the usefulness of the proposed WSNM, we apply it to two typical low level vision applications: image denoising and background subtraction. For image denoising, similar to WNNM \cite{wnnm}, WSNM is applied to the matrix of image nonlocal similar patches for noise removal. For background modeling, like RPCA \cite{RPCA}, we propose a WSNM based RPCA method to capture the ``low rank + sparse'' structure of input data matrix.

\subsection{WSNM in Image Denoising}\label{proposed_denoising_method}
For a local patch $\mathbf{y}_i$ in a degraded image $\mathbf{y}$, we search its nonlocal similar patches $\{\tilde{\mathbf{y}}_i\}_{i=1}^{n_i}$ by the block matching method proposed in \cite{BM3D}. Then, $\{\tilde{\mathbf{y}}_i\}_{i=1}^{n_i}$ is stacked into a matrix $\mathbf{Y}_i$, whose columns are composed of the vectorized patches $\tilde{\mathbf{y}}_i$ $(i=1, \ldots, n_i)$. According to degradation model of additive white Gaussian noise, we have $\mathbf{Y}_i = \mathbf{X}_i + \mathbf{N}_i$, where $\mathbf{X}_i$ and $\mathbf{N}_i$ are the patch matrices of original image and noise, respectively. Under the assumption of low rank, the matrix $\mathbf{X}_i$ can be estimated from $\mathbf{Y}_i$ by using the LRMA methods. Hence, we apply the proposed WSNM model to estimate $\mathbf{X}_i$, and its corresponding optimization problem can be defined as
\begin{equation}\label{patch-model}
    \hat{\mathbf{X}}_i = \arg \min_{\mathbf{X}_i} \frac{1}{\sigma_{n}^{2}} \| \mathbf{Y}_i -\mathbf{X}_i \|_{F}^{2}
                         + \| \mathbf{X}_i \|_{w, S_p}^{p},
\end{equation}
where $\sigma_{n}^{2}$ denotes the noise variance, the first term of (\ref{patch-model}) represents the F-norm data fidelity term, and the second term plays the role of low rank regularization. Usually, $\sigma_j(\mathbf{X}_i)$, the $j$-th singular value of $\mathbf{X}_i$, with a larger value is more important than small ones since it represents the energy of the $j$-th component of $\mathbf{X}_i$. Similarly, $\delta_j(\hat{\mathbf{X}}_i)$, the $j$-th singular value of the optimal solution of model (\ref{patch-model}), owns the same property such that the larger the value of $\delta_j(\hat{\mathbf{X}}_i)$, the less it should be shrunk. Therefore, an intuitive way for setting weights is that the weight should be inversely proportional to $\delta_j(\hat{\mathbf{X}}_i)$, and we let
\begin{equation}\label{weight-setting}
    w_j = c\sqrt{n}\bigg/(\delta_j^{1/p}(\hat{\mathbf{X}}_i) + \varepsilon),
\end{equation}
where $n$ is the number of similar patches in $\mathbf{Y}_i$, $\varepsilon$ is set to $10^{-16}$ to avoid dividing by zero, and $c = 2\sqrt{2}\sigma^2_n$. Since $\delta_j(\hat{\mathbf{X}}_i)$ is unavailable before $\hat{\mathbf{X}}$ is estimated, it can be initialized by
\begin{equation}\label{weight-initial}
    \delta_j(\hat{\mathbf{X}}_i) = \sqrt{\max\{\sigma^2_j(\mathbf{Y}_i) - n\sigma^2_n, 0\}}.
\end{equation}

To restore clean image iteratively, we adopt the iterative regularization scheme in \cite{SAIST}, which adds filtered residual back to the denoised image as follows:
\begin{equation}\label{update-rule1}
    \mathbf{y}^{(k)} = \hat{\mathbf{x}}^{(k-1)} + \alpha (\mathbf{y} - \hat{\mathbf{x}}^{(k-1)}),
\end{equation}
where $k$ denotes the iteration number and $\alpha$ is a relaxation parameter. Finally, by aggregating all the denoised patches together, the image $\mathbf{x}$ can be reconstructed. The whole denoising algorithm is summarized in Algorithm \ref{propose_alg}.

\SetAlFnt{\footnotesize}{ 
\begin{algorithm}[]\label{propose_alg}
\SetAlgoLined
\caption{Image Denoising by WSNM}
\KwIn{Noisy image $\mathbf{y}$}
\KwOut{Denoised image $\hat{\mathbf{x}}^{K}$\\}
\BlankLine
Initialization:\\
\qquad Initialize $\hat{\mathbf{x}}^{0} = \mathbf{y}, \hat{\mathbf{y}}^{0} = \hat{\mathbf{y}}$\\
\For{$k = 1:K$}
{
    Iterative regularization $\mathbf{y}^{(k)} = \hat{\mathbf{x}}^{(k-1)} + \alpha (\mathbf{y} - \mathbf{x}^{(k-1)})$\;
    \For{each patch $\mathbf{y}_{j}^{k}$}
    {
        Find similar patches to form matrix $\mathbf{Y}_{j}$\;
        Estimate weight vector $\mathbf{w}$ by Eq.(\ref{weight-setting})\;
        Singular value decomposition $[\mathbf{U}, \mathbf{\Sigma}, \mathbf{V}] = SVD(\mathbf{Y}_j)$\;
        Calculate $\mathbf{\Delta}$ by using Eq.(\ref{equivalence-problem-decoupled})\;
        Get the estimation: $\hat{\mathbf{X}}_j = \mathbf{U} \mathbf{\Delta} \mathbf{V}^{T}$\;
    }
    Aggregate $\mathbf{X}_j$ to form the denoised image $\mathbf{x}^{k}$\;
}
\textbf{Return} The final denoised image $\hat{\mathbf{x}}^{K}$\;
\end{algorithm}}

\subsection{WSNM-RPCA for Background Subtraction}\label{WSNM-RPCA-Background}

\SetAlFnt{\footnotesize}{ 
\begin{algorithm}[]\label{propose-wsnm-rpca}
\SetAlgoLined
\caption{WSNM-RPCA}
\KwIn{Observed data $\mathbf{Y}$, weight vector $w$, power $p$}
\KwOut{Matrices $\mathbf{X}$ and $\mathbf{E}$\\}
\BlankLine
Initialization:\\
\qquad $\mu_0 > 0, \rho >1, k = 0, \mathbf{X} = \mathbf{Y}, \mathbf{Z=0}$\\
\While{not convergence}
{
    $\mathbf{E}_{k+1} = \argmin_{E} \|\mathbf{E}\|_{1} + \frac{\mu_k}{2} \|\mathbf{Y} + \mu_{k}^{-1}\mathbf{Z_k} - \mathbf{X}_{k} - \mathbf{E}\|_{F}^{2}$\;
    $\mathbf{X}_{k+1} = \argmin_{X} \|\mathbf{X}\|_{w,S_p}^{p} + \frac{\mu_k}{2} \|\mathbf{Y} + \mu_{k}^{-1}\mathbf{Z_k} - \mathbf{E}_{k+1} - \mathbf{X}\|_{F}^{2}$\;
    $\mathbf{Z}_{k+1} = \mathbf{Z}_{k} + \mu_{k}(\mathbf{Y} - \mathbf{X}_{k+1} - \mathbf{E}_{k+1})$\;
    $\mu_{k+1} = \rho \ast \mu_{k}$\;
    $k = k + 1$\;
}
\textbf{Return} $\mathbf{X}$ and $\mathbf{E}$\;
\end{algorithm}}

Background subtraction from video sequences captured by a static camera can be considered as a low-rank matrix analysis problem \cite{RPCA,rpca-acm}, which can be formulated by the following NNM based RPCA (NNM-RPCA) model \cite{alm}:
\begin{equation}\label{rpca-model}
    \min_{\mathbf{E},\mathbf{X}} \|\mathbf{E}\|_{1} + \|\mathbf{X}\|_{\ast}, {s.t. } \mathbf{Y} = \mathbf{X} + \mathbf{E},
\end{equation}
where the error matrix $\mathbf{E}$ is enforced by $l_1$-norm so that the model is robust to sparse noise. Here, we propose a new RPCA model namely WSNM-RPCA by using WSNM to replace the NNM in (\ref{rpca-model}):
\begin{equation}\label{wsnm-rpca}
    \min_{\mathbf{E},\mathbf{X}} \|\mathbf{E}\|_{1} + \|\mathbf{X}\|_{w, S_p}^{p}, {s.t. } \mathbf{Y} = \mathbf{X} + \mathbf{E}.
\end{equation}
Its augmented Lagrange function is
\begin{equation}\label{wsnm-rpca-fun}
    \begin{aligned}
    &\mathbf{L}(\mathbf{X},\mathbf{E},\mathbf{Z},\mu) = \\
    &\|\mathbf{E}\|_{1} + \|\mathbf{X}\|_{w, S_p}^{p} + \langle \mathbf{Z}, \mathbf{Y} - \mathbf{X} - \mathbf{E} \rangle + \frac{\mu}{2}\|\mathbf{Y} - \mathbf{X} - \mathbf{E}\|_{F}^{2},
    \end{aligned}
\end{equation}
where $\mathbf{Z}$ is the Lagrange multiplier, $\mu$ is a positive scalar, and the weights are set as:
\begin{equation}\label{wsnm-rpca-weight}
    w_i = C\sqrt{mn}\bigg/(\sigma_i(\mathbf{Y}) + \varepsilon).
\end{equation}
Minimizing Eq.(\ref{wsnm-rpca-fun}) directly is still rather challenging. According to the recently developed alternating direction minimization technique \cite{alm}, the Lagrangian function can be solved by minimizing each variable alternatively while fixing the other variables. The optimization procedure is described in Algorithm \ref{propose-wsnm-rpca}.

Here, we will briefly analyze the convergence of the proposed algorithm for WSNM-RPCA. Since the weighted schatten $p$-norm is not convex and has no general form of subgradient, the convergence analysis of Algorithm \ref{propose-wsnm-rpca} is much more difficult. The following theorem gives the convergence analysis:
\begin{theorem}\label{theorem-wsnm-rpca}
  If the weights satisfy $0\leq w_1 \leq w_2 \leq \ldots \leq w_r$, the sequence $\mathbf{E}_k$ and $\mathbf{X}_k$ obtained by Algorithm 4 (WSNM-RPCA) has at least one accumulation point, and the point sequence satisfy:
  \begin{equation}\label{}
    \lim_{k\rightarrow\infty} \|\mathbf{E}_{k+1} - \mathbf{E}_k\|_{F}^{2} + \|\mathbf{X}_{k+1} - \mathbf{X}_k\|_{F}^{2} = 0.
  \end{equation}
  Moreover, the accumulation points form a compact continuum C.
\end{theorem}
\begin{pf}
  The proof can be found in the appendix.
\end{pf}
Although Theorem \ref{theorem-wsnm-rpca} only ensures that the change of the variables in consecutive iterations tends to zero, there is no guarantee that the sequence will converge to a local minimum. However, in our experiments the proposed method converges fast, which confirms the feasibility the proposed optimization.

\begin{figure}[H]
\setlength{\abovecaptionskip}{0pt}  
\setlength{\belowcaptionskip}{0pt} 
\renewcommand{\figurename}{Figure}
\centering
\includegraphics[width=0.5\textwidth]{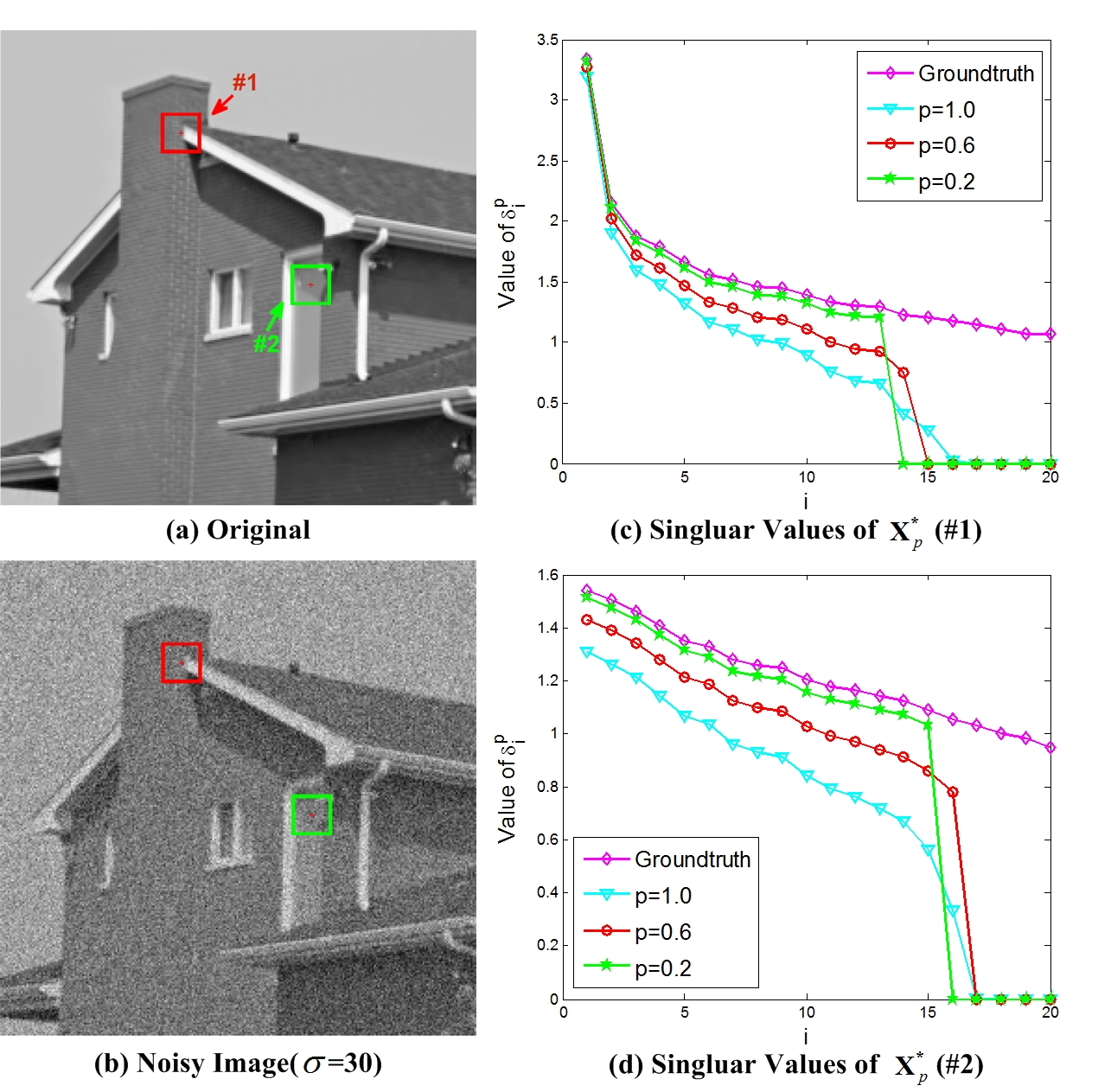}
\caption{Illustration of the over-shrinkage problem.}
\label{fig:suppression-p}
\end{figure}

\section{Experimental Results and Analysis}\label{experiment}

\subsection{Image Denoising}\label{experiment-denoising}

Firstly, we test the performance of the proposed WSNM in image denoising, and compare it with six representative algorithms: block-matching 3D filtering \cite{BM3D} (\textbf{BM3D}), patch-based near-optimal image denoising \cite{PBNO} (\textbf{PBNO}), spatially adaptive iterative singular-value thresholding \cite{SAIST} (\textbf{SAIST}), expected patch log likelihood for image denoising \cite{EPLL} (\textbf{EPLL}), global image denoising \cite{GID} (\textbf{GID}), and weighted nuclear norm minimization \cite{wnnm} (\textbf{WNNM}). It is worth to note that those methods, especially WNNM, are the schemes in the open literature whose performance has shown convincing improvements over BM3D. Therefore, it is significative to compare with those algorithms. The denoising results of all methods are generated from the source codes or executables provided by their authors, and we keep the parameter settings mentioned in original papers for all the test images. The code and data of the proposed method are available on the website \textbf{https://sites.google.com/site/yuanxiehomepage/}.

Several parameters need to be set in the proposed algorithm. According to the analysis of power $p$ (discussed in Section \ref{experiment-p-norm}), we choose $p = \{1.0, 0.85, 0.75, 0.7, 0.1, 0.05\}$ for $\sigma_n = 20$, $30$, $50$, $60$, $75$, and $100$ in proposed WSNM, respectively. Other parameters settings are the same as WNNM. All experiments are implemented in Matlab on a PC with $3.5$GHz CPU and $16$GB RAM. 

\begin{figure}[H]
\setlength{\abovecaptionskip}{0pt}  
\setlength{\belowcaptionskip}{0pt} 
\renewcommand{\figurename}{Figure}
\centering
\includegraphics[width=0.5\textwidth]{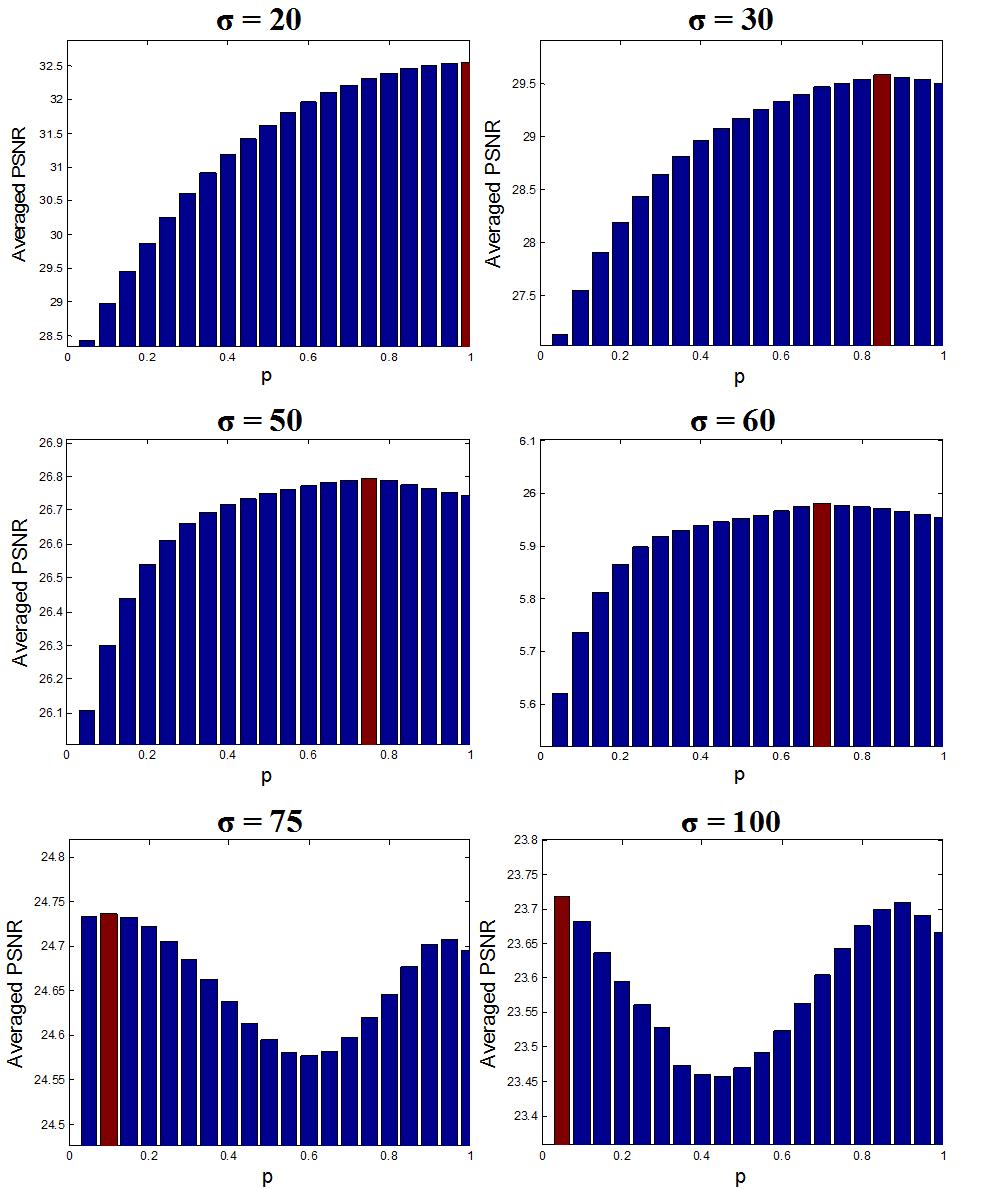}
\caption{The influence of changing $p$ upon denoised results under different noise levels on 40 images randomly selected from the Berkeley Segmentation Dataset \cite{BSDS}.}
\label{fig:p-change}
\end{figure}

\subsubsection{\textbf{Advantages of the Weighted Schatten $p$-Norm}}\label{experiment-p-norm}

This subsection illustrates the advantages of the proposed weighted schatten $p$-norm. Here, we use a test to point out that weighted nuclear norm suffers from a problem: the obtained singular values can be over-shrunk, which leads to solution deviation. In Fig.\ref{fig:suppression-p}, we use both WSNM and WNNM to perform low rank approximation on the two patches (marked by the red and green boxes) randomly cropped from the noisy image (Fig.\ref{fig:suppression-p} (b)). Let $\{\delta_{i}\}$ be the singular values of the matrix of similar patches in the clean image Fig.\ref{fig:suppression-p} (a), and $\{\delta^{(p)}_{i}\}$ be the singular values of the optimal solution $\mathbf{X}^{*}_{p}$ of model (\ref{patch-model}). We show the solution $\{\delta^{(p)}_{i}\}$ in Fig.\ref{fig:suppression-p} (c) and Fig.\ref{fig:suppression-p} (d) for patches \#1 and \#2, respectively. From Fig.\ref{fig:suppression-p} (c), we can see that the $\{\delta^{(p=1)}_{i}\}$ (denoted by cyan line) are deviated far from $\{\delta_{i}\}$ (denoted by magenta line), meaning that the {\it over-shrinkage} is serious. As $p$ decreases, more high rank components of $\{\delta^{p}_{i}\}$ become zeros, while the low rank components of $\{\delta^{p}_{i}\}$ are getting closer to $\{\delta_{i}\}$.

In addition, it is necessary for us to analyze the suitable setting of power $p$ for each noise level $\sigma_n$. So, we randomly select $40$ images from the Berkeley Segmentation Dataset \cite{BSDS}, add noise to them, and test the proposed WSNM with different power $p$ under different noise levels. In each subfigure of Fig.\ref{fig:p-change}, horizontal coordinate denotes the values of power $p$ changing from $0.05$ to $1$ with interval $0.05$, vertical coordinate represents the averaged value of PSNR under certain noise level. In this test, six noise levels $\sigma_n = \{20, 30, 50, 60, 75, 100\}$ are used. When handling low and medium noise levels ($20$, $30$ and $50$), as shown in first three subfigures of Fig. \ref{fig:p-change}, the best values of $p$ are $1.0$, $0.85$ and $0.75$, respectively. With the noise level becoming stronger, more rank components of data are contaminated. Consequently, the high rank parts will be penalized heavily, which means that the small values of $p$ are preferred ($0.7$, $0.1$ and $0.05$ for noise levels $60$, $75$ and $100$, respectively), as demonstrated in last three subfigures of Fig. \ref{fig:p-change}. To sum up, the best value of power $p$ is inversely proportional to the noise level, and the empirical values will be directly applied to the test images in the next subsection.


\subsubsection{\textbf{Comparison with State-of-the-Art Methods}}\label{comparison-experiment}

\begin{table*}[htbp]
\centering
\renewcommand{\arraystretch}{1}
\caption{Denoising results (PSNR) by different methods}\label{comparison-table}
\vspace{4pt}
\footnotesize
\begin{tabular}{IcIc|c|c|c|c|c|cIc|c|c|c|c|c|cI}
\whline
\multirow{2}{*}{} &
\multicolumn{7}{cI}{$\sigma_n = 20$} &
\multicolumn{7}{cI}{$\sigma_n = 30$} \\

\cline{2-15}   
         & BM3D & PBNO & EPLL & GID & SAIST & WNNM & WSNM          & BM3D & PBNO & EPLL & GID & SAIST & WNNM & WSNM\\
\whline
C.Man    & 30.48 & 29.61 & 30.34 & 29.31 & 30.45 & 30.75 & \textbf{30.77}    & 28.63 & 27.87 & 28.36 & 27.84 & 27.47 & 28.80 & \textbf{28.83}\\
\hline
House    & 33.77 & 33.58 & 32.98 & 32.81 & 33.75 & 34.01 & \textbf{34.05}    & 32.08 & 31.92 & 31.22 & 30.35 & 31.39 & 32.52 & \textbf{32.54}\\
\hline
Peppers  & 31.29 & 30.55 & 31.17 & 30.17 & 31.32 & 31.53 & \textbf{31.55}    & 29.28 & 28.81 & 29.16 & 28.16 & 28.33 & 29.48 & \textbf{29.54}\\
\hline
Monarch  & 30.35 & 29.55 & 30.48 & 29.65 & 30.76 & 31.10 & \textbf{31.13}    & 28.36 & 27.85 & 28.35 & 27.60 & 28.03 & 28.91 & \textbf{28.94}\\
\hline
Airplane & 32.53 & 32.06 & 32.41 & 31.48 & 32.39 & \textbf{32.82} & \textbf{32.82}    & 27.56 & 30.21 & 30.41 & 29.47 & 29.35 & 30.87 & \textbf{30.92}\\
\hline
Barbara  & 31.77 & 31.06 & 29.76 & 30.21 & 32.10 & 32.19 & \textbf{32.20}    & 29.81 & 29.50 & 27.56 & 27.95 & 30.04 & 30.31 & \textbf{30.38}\\
\hline
Boat     & 30.88 & 30.39 & 30.66 & 29.53 & 30.84 & 31.00 & \textbf{31.02}    & 29.11 & 28.81 & 28.89 & 27.66 & 28.83 & 29.24 & \textbf{29.26}\\
\hline
Bridge   & 27.27 & 26.70 & 27.49 & 26.49 & 27.31 & \textbf{27.42} & \textbf{27.42}    & 25.46 & 25.22 & 25.68 & 24.78 & 25.43 & 25.62 & \textbf{25.66}\\
\hline
Couple   & 30.76 & 30.22 & 30.54 & 29.28 & 30.66 & \textbf{30.82} & \textbf{30.82}    & 28.86 & 28.58 & 28.61 & 27.15 & 28.58 & 28.98 & \textbf{29.02}\\
\hline
F.print  & 28.80 & 27.76 & 28.28 & 27.95 & 28.99 & 29.02 & \textbf{29.04}    & 26.82 & 26.35 & 26.18 & 26.00 & 26.82 & 26.99 & \textbf{27.10}\\
\hline
F.stones & 29.57 & 28.78 & 29.17 & 29.28 & 29.61 & 29.99 & \textbf{30.01}    & 27.81 & 27.22 & 27.38 & 27.31 & 27.98 & 28.20 & \textbf{28.32}\\
\hline
Lolly     & 31.48 & 31.57 & 31.24 & 30.81 & 31.40 & \textbf{31.44} & \textbf{31.44}    & 30.44 & 30.49 & 30.15 & 29.67 & 30.35 & 30.46 & \textbf{30.50}\\
\hline
Hill     & 30.72 & 30.32 & 30.49 & 29.59 & 30.58 & 30.81 & \textbf{30.83}    & 29.15 & 28.95 & 28.90 & 27.75 & 28.94 & 29.25 & \textbf{29.27}\\
\hline
J.Bean   & 35.64 & 35.22 & 35.13 & 34.48 & 36.01 & \textbf{36.18} & \textbf{36.18}    & 33.39 & 33.11 & 32.79 & 32.22 & 31.40 & 33.79 & \textbf{33.88}\\
\hline
Lena     & 33.05 & 32.75 & 32.61 & 31.74 & 33.08 & 33.12 & \textbf{33.13}    & 31.26 & 31.16 & 30.78 & 29.83 & 30.77 & 31.43 & \textbf{31.48}\\
\hline
Man      & 30.59 & 30.15 & 30.63 & 29.59 & 30.54 & 30.74 & \textbf{30.77}    & 28.86 & 28.65 & 28.82 & 27.82 & 28.68 & \textbf{29.00} & \textbf{29.00}\\
\hline
Parrot   & 29.96 & 29.22 & 29.97 & 28.96 & 29.97 & 30.19 & \textbf{30.21}    & 28.11 & 27.46 & 28.07 & 27.52 & 27.62 & \textbf{28.33} & \textbf{28.33}\\
\hline
Rice     & 34.60 & 34.49 & 33.59 & 34.62 & 34.73 & \textbf{35.25} & \textbf{35.25}    & 32.24 & 31.93 & 31.12 & 32.17 & 32.22 & 32.74 & \textbf{33.00}\\
\hline
Straw    & 27.07 & 25.86 & 26.92 & 26.63 & 27.23 & \textbf{27.44} & \textbf{27.44}    & 24.94 & 24.70 & 24.74 & 24.59 & 24.74 & 25.23 & \textbf{25.48}\\
\hline
Truck    & 30.95 & 30.77 & 30.97 & 29.87 & 30.77 & 31.03 & \textbf{31.06}    & 29.54 & 29.39 & 29.46 & 28.31 & 29.12 & 29.55 & \textbf{29.63}\\
\whline
AVG.     & 31.08 & 30.54 & 30.74 & 30.12 & 31.12 & 31.34 & \textbf{31.37}    & 29.09 & 28.91 & 28.83 & 28.21 & 28.81 & 29.48 & \textbf{29.56}\\
\whline

\multirow{2}{*}{} &
\multicolumn{7}{cI}{$\sigma_n = 50$} &
\multicolumn{7}{cI}{$\sigma_n = 60$} \\
\cline{2-15}   
         & BM3D & PBNO & EPLL & GID & SAIST & WNNM & WSNM          & BM3D & PBNO & EPLL & GID & SAIST & WNNM & WSNM\\
\whline
C.Man    & 26.13 & 25.71 & 26.02 & 25.48 & 25.94 & 26.42 & \textbf{26.44}    & 25.31 & 24.98 & 25.20 & 24.50 & 25.15 & 25.54 & \textbf{25.67}\\
\hline
House    & 29.69 & 29.44 & 28.76 & 27.62 & 29.99 & 30.23 & \textbf{30.36}    & 28.73 & 28.62 & 27.84 & 26.66 & 28.88 & 29.37 & \textbf{29.52}\\
\hline
Peppers  & 26.68 & 26.46 & 26.62 & 25.60 & 26.60 & 26.81 & \textbf{26.94}    & 25.81 & 25.66 & 25.67 & 24.64 & 25.63 & 25.98 & \textbf{26.06}\\
\hline
Monarch  & 25.81 & 25.53 & 25.77 & 24.97 & 26.09 & 26.18 & \textbf{26.30}    & 24.97 & 24.64 & 24.85 & 24.15 & 24.94 & 25.33 & \textbf{25.48}\\
\hline
Airplane & 25.10 & 27.77 & 27.88 & 26.91 & 28.25 & 28.44 & \textbf{28.49}    & 27.32 & 26.98 & 26.97 & 25.82 & 26.64 & 27.60 & \textbf{27.64}\\
\hline
Barbara  & 27.22 & 26.95 & 24.82 & 25.17 & 27.49 & 27.79 & \textbf{27.83}    & 26.28 & 26.08 & 23.87 & 24.19 & 26.40 & \textbf{26.97} & 26.84\\
\hline
Boat     & 26.78 & 26.67 & 26.65 & 25.59 & 26.63 & 26.97 & \textbf{27.01}    & 26.02 & 25.94 & 25.84 & 24.68 & 25.52 & \textbf{26.21} & \textbf{26.21}\\
\hline
Bridge   & 23.57 & 23.49 & 23.69 & 22.88 & 23.49 & 23.73 & \textbf{23.77}    & 23.02 & 22.90 & 23.08 & 22.19 & 22.85 & 23.13 & \textbf{23.16}\\
\hline
Couple   & 26.46 & 26.30 & 26.23 & 24.64 & 26.29 & 26.65 & \textbf{26.71}    & 25.66 & 25.43 & 25.40 & 24.01 & 24.98 & 25.87 & \textbf{25.89}\\
\hline
F.print  & 24.52 & 24.29 & 23.59 & 23.09 & 24.54 & 24.67 & \textbf{24.73}    & 23.75 & 23.57 & 22.65 & 21.90 & 23.71 & \textbf{23.95} & 23.91\\
\hline
F.stones & 25.10 & 24.86 & 24.89 & 24.26 & 25.41 & 25.41 & \textbf{25.63}    & 24.13 & 23.82 & 23.93 & 23.11 & 23.88 & 24.47 & \textbf{24.74}\\
\hline
Lolly     & 28.94 & 28.72 & 28.52 & 28.20 & 28.82 & 28.95 & \textbf{29.00}    & 28.29 & 27.98 & 27.87 & 27.53 & 28.02 & 28.38 & \textbf{28.40}\\
\hline
Hill     & 27.19 & 27.02 & 26.95 & 25.93 & 27.04 & 27.34 & \textbf{27.36}    & 26.52 & 26.27 & 26.27 & 25.32 & 26.39 & \textbf{26.72} & 26.67\\
\hline
J.Bean   & 30.66 & 30.32 & 29.92 & 30.01 & 29.70 & 30.78 & \textbf{31.08}    & 29.73 & 29.23 & 28.93 & 29.01 & 29.29 & 30.10 & \textbf{30.15}\\
\hline
Lena     & 29.05 & 28.81 & 28.42 & 27.69 & 29.01 & 29.24 & \textbf{29.28}    & 28.27 & 27.92 & 27.59 & 26.91 & 28.00 & 28.54 & \textbf{28.59}\\
\hline
Man      & 26.80 & 26.72 & 26.72 & 25.83 & 26.67 & 26.93 & \textbf{26.98}    & 26.13 & 26.00 & 26.00 & 25.14 & 25.78 & \textbf{26.24} & \textbf{26.24}\\
\hline
Parrot   & 25.89 & 25.37 & 25.83 & 25.33 & 25.76 & 26.00 & \textbf{26.10}    & 25.14 & 24.69 & 25.03 & 24.41 & 24.95 & 25.26 & \textbf{25.36}\\
\hline
Rice     & 29.18 & 28.64 & 28.03 & 28.49 & 29.43 & 29.65 & \textbf{29.89}    & 28.05 & 27.60 & 26.93 & 26.62 & 28.30 & 28.68 & \textbf{28.79}\\
\hline
Straw    & 22.40 & 22.81 & 22.00 & 21.98 & 22.65 & 22.74 & \textbf{22.93}    & 21.63 & 22.01 & 21.06 & 20.93 & 22.13 & \textbf{22.20} & 22.04\\
\hline
Truck    & 27.82 & 27.51 & 27.63 & 26.85 & 27.52 & 27.85 & \textbf{27.92}    & 27.21 & 26.97 & 27.00 & 26.32 & 27.03 & \textbf{27.21} & 27.11\\
\whline
AVG.     & 26.75 & 26.67 & 26.45 & 25.83 & 26.86 & 27.14 & \textbf{27.24}    & 26.09 & 25.86 & 25.59 & 24.90 & 25.92 & 26.38 & \textbf{26.42}\\
\whline

\multirow{2}{*}{} &
\multicolumn{7}{cI}{$\sigma_n = 75$} &
\multicolumn{7}{cI}{$\sigma_n = 100$} \\
\cline{2-15}   
         & BM3D & PBNO & EPLL & GID & SAIST & WNNM & WSNM          & BM3D & PBNO & EPLL & GID & SAIST & WNNM & WSNM\\
\whline
C.Man    & 24.32 & 24.01 & 24.19 & 23.26 & 24.27 & 24.55 & \textbf{24.60}    & 23.08 & 22.65 & 22.85 & 21.72 & 23.08 & 23.36 & \textbf{23.40}\\
\hline
House    & 27.50 & 27.15 & 26.68 & 25.16 & 27.90 & 28.25 & \textbf{28.40}    & 25.87 & 25.42 & 25.19 & 23.59 & 26.45 & 26.67 & \textbf{26.80}\\
\hline
Peppers  & 24.73 & 24.55 & 24.56 & 23.34 & 24.68 & \textbf{24.92} & 24.88    & 23.39 & 23.03 & 23.08 & 21.61 & 23.35 & 23.46 & \textbf{23.53}\\
\hline
Monarch  & 23.90 & 23.62 & 23.71 & 22.77 & 23.95 & 24.31 & \textbf{24.37}    & 22.51 & 22.19 & 22.23 & 20.83 & 22.63 & 22.95 & \textbf{23.00}\\
\hline
Airplane & 26.31 & 25.83 & 25.83 & 24.69 & 25.82 & 26.58 & \textbf{26.60}    & 22.11 & 24.31 & 24.35 & 23.28 & 24.55 & 25.23 & \textbf{25.27}\\
\hline
Barbara  & 25.12 & 24.94 & 22.94 & 23.06 & 25.35 & 25.81 & \textbf{25.85}    & 23.62 & 23.42 & 22.14 & 21.76 & 23.98 & 24.37 & \textbf{24.40}\\
\hline
Boat     & 25.14 & 24.85 & 24.88 & 23.81 & 24.80 & \textbf{25.29} & 25.25    & 23.97 & 23.62 & 23.71 & 22.74 & 23.67 & 24.10 & \textbf{24.16}\\
\hline
Bridge   & 22.40 & 22.26 & 22.39 & 21.52 & 22.07 & 22.43 & \textbf{22.50}    & 21.60 & 21.42 & 21.58 & 20.74 & 21.21 & 21.58 & \textbf{21.66}\\
\hline
Couple   & 24.70 & 24.51 & 24.44 & 23.27 & 24.17 & 24.85 & \textbf{24.89}    & 23.51 & 23.28 & 23.32 & 22.38 & 23.01 & 23.55 & \textbf{23.62}\\
\hline
F.print  & 22.83 & 22.67 & 21.46 & 20.43 & 22.72 & \textbf{23.02} & 23.00    & 21.61 & 21.50 & 19.84 & 18.74 & 21.51 & 21.81 & \textbf{21.87}\\
\hline
F.stones & 22.92 & 22.71 & 22.69 & 21.47 & 22.79 & 23.28 & \textbf{23.35}    & 21.31 & 21.07 & 21.03 & 19.44 & 21.42 & 21.63 & \textbf{21.82}\\
\hline
Lolly     & 27.41 & 27.24 & 27.02 & 26.73 & 27.25 & 27.52 & \textbf{27.57}    & 26.21 & 25.98 & 25.89 & 25.54 & 25.98 & 26.24 & \textbf{26.40}\\
\hline
Hill     & 25.67 & 25.45 & 25.45 & 24.62 & 25.50 & 25.87 & \textbf{25.90}    & 24.58 & 24.33 & 24.42 & 23.79 & 24.29 & 24.75 & \textbf{24.83}\\
\hline
J.Bean   & 28.70 & 27.84 & 27.74 & 27.95 & 28.40 & 29.00 & \textbf{29.02}    & 27.29 & 26.29 & 26.34 & 26.08 & 27.15 & 27.52 & \textbf{27.80}\\
\hline
Lena     & 27.25 & 27.00 & 26.57 & 25.96 & 26.97 & \textbf{27.54} & 27.52    & 25.95 & 25.60 & 25.30 & 24.64 & 25.81 & 26.20 & \textbf{26.31}\\
\hline
Man      & 25.31 & 25.11 & 25.14 & 24.38 & 25.06 & 25.42 & \textbf{25.48}    & 24.22 & 23.98 & 24.07 & 23.33 & 23.98 & 24.35 & \textbf{24.41}\\
\hline
Parrot   & 24.18 & 23.69 & 24.03 & 23.54 & 24.11 & 24.32 & \textbf{24.36}    & 22.95 & 22.49 & 22.70 & 21.93 & 23.00 & 23.19 & \textbf{23.20}\\
\hline
Rice     & 26.65 & 26.18 & 25.61 & 24.30 & 27.08 & 27.28 & \textbf{27.51}    & 24.73 & 24.22 & 23.90 & 21.90 & 25.36 & 25.39 & \textbf{25.54}\\
\hline
Straw    & 20.72 & 21.04 & 20.07 & 19.55 & 21.08 & 21.12 & \textbf{21.18}    & 19.58 & 19.86 & 19.01 & 18.41 & 19.54 & 19.67 & \textbf{19.90}\\
\hline
Truck    & 26.51 & 26.22 & 26.26 & 25.59 & 26.28 & 26.47 & \textbf{26.52}    & 25.36 & 25.30 & 25.33 & 23.09 & 25.20 & 25.47 & \textbf{25.79}\\
\whline
AVG.     & 25.11 & 24.84 & 24.58 & 23.77 & 25.01 & 25.39 & \textbf{25.44}    & 23.67 & 23.49 & 23.31 & 22.28 & 23.75 & 24.07 & \textbf{24.19}\\
\whline

\end{tabular}
\end{table*}

\begin{figure}[htbp]
\setlength{\abovecaptionskip}{0pt}  
\setlength{\belowcaptionskip}{0pt} 
\renewcommand{\figurename}{Figure}
\centering
\includegraphics[width=0.5\textwidth]{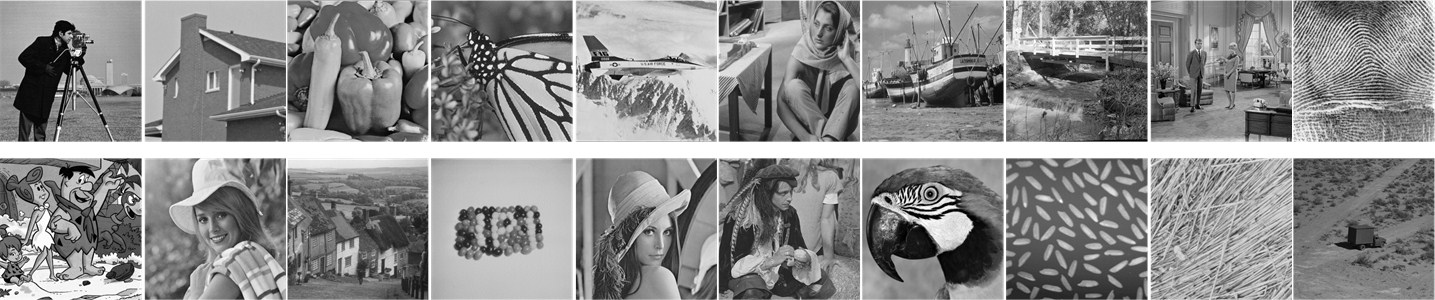}
\caption{The 20 test images for image denoising.}
\label{fig:dataset}
\end{figure}

\begin{figure*}[htbp]
\setlength{\abovecaptionskip}{0pt}  
\setlength{\belowcaptionskip}{0pt} 
\renewcommand{\figurename}{Figure}
\centering
\includegraphics[width=0.8\textwidth]{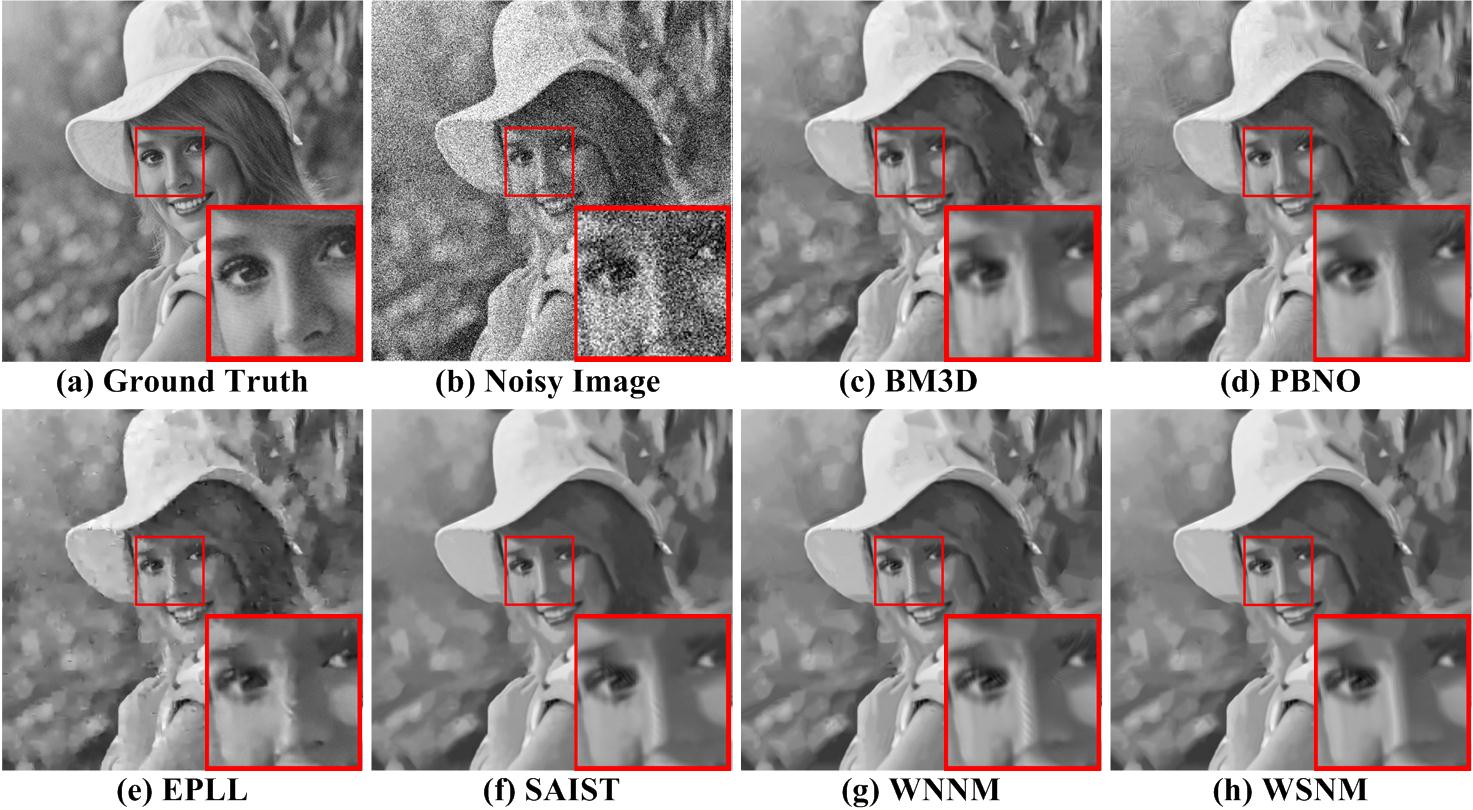}
\caption{Denoising results on image {\it Lolly} by different methods (noise level $\sigma_n = 50$). (a) Ground Truth. (b) Noisy Image. (c) BM3D, PSNR = 28.94dB. (d) PBNO, PSNR = 28.72dB. (e) EPLL, PSNR = 28.52dB. (f) SAIST, PSNR = 28.82dB. (g) WNNM, PSNR = 28.95dB. (h) WSNM, PSNR = 29.00dB. The figure is better viewed in zoomed PDF.}
\label{fig:compare-girl-sigma50}
\end{figure*}
\begin{figure*}[htbp]
\setlength{\abovecaptionskip}{0pt}  
\setlength{\belowcaptionskip}{0pt} 
\renewcommand{\figurename}{Figure}
\centering
\includegraphics[width=0.8\textwidth]{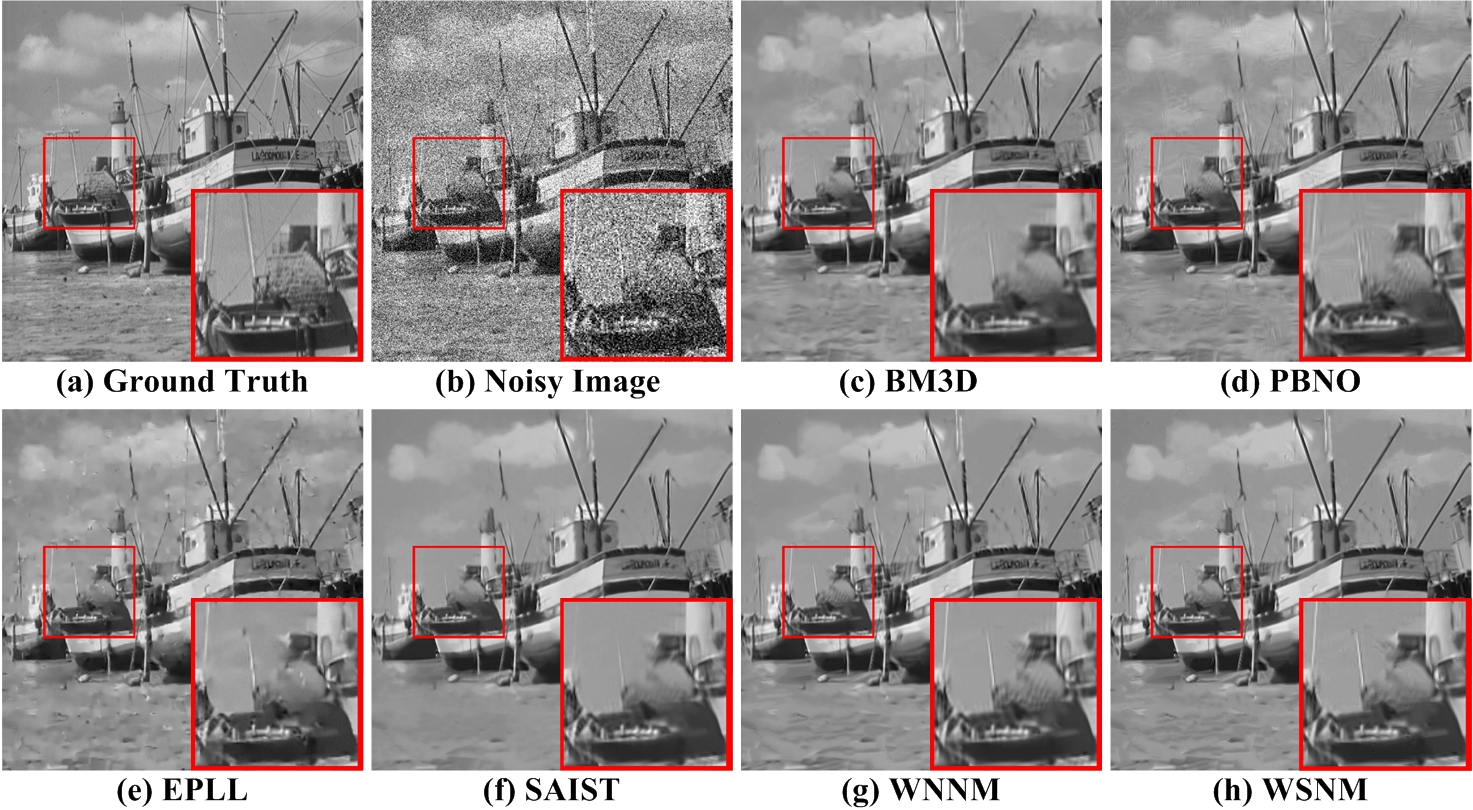}
\caption{Denoising results on image {\it Boat} at noise level $\sigma_n = 50$. (a) Ground Truth. (b) Noisy Image. (c) BM3D, PSNR = 26.78dB. (d) PBNO, PSNR = 26.67dB. (e) EPLL, PSNR = 26.65dB. (f) SAIST, PSNR = 26.63dB. (g) WNNM, PSNR = 26.97dB. (h) WSNM, PSNR = 27.01dB.}
\label{fig:compare-parrot-sigma50}
\end{figure*}
\begin{figure*}[htbp]
\setlength{\abovecaptionskip}{0pt}  
\setlength{\belowcaptionskip}{0pt} 
\renewcommand{\figurename}{Figure}
\centering
\includegraphics[width=0.8\textwidth]{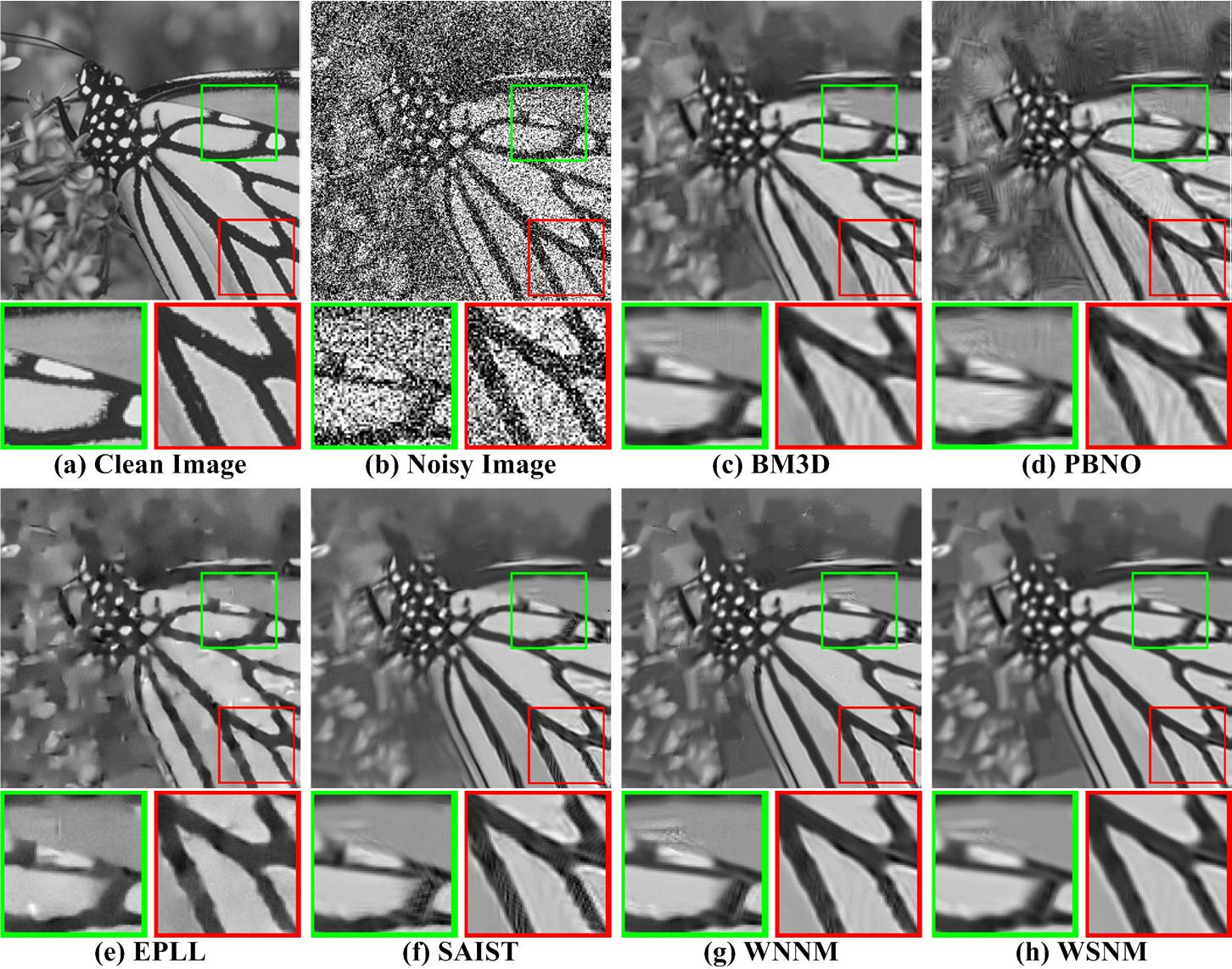}
\caption{Denoising results on image {\it Monarch} by different methods (noise level $\sigma_n = 100$). (a) Ground Truth. (b) Noisy Image. (c) BM3D, PSNR = 22.51dB. (d) PBNO, PSNR = 22.19dB. (e) EPLL, PSNR = 22.23dB. (f) SAIST, PSNR = 22.63dB. (g) WNNM, PSNR = 22.95dB. (h) WSNM, PSNR = 23.00dB. The figure is better viewed in zoomed PDF.}
\label{fig:compare-monarch-sigma100}
\end{figure*}

\begin{figure*}[!ht]
\setlength{\abovecaptionskip}{0pt}  
\setlength{\belowcaptionskip}{0pt} 
\renewcommand{\figurename}{Figure}
\centering
\includegraphics[width=0.8\textwidth]{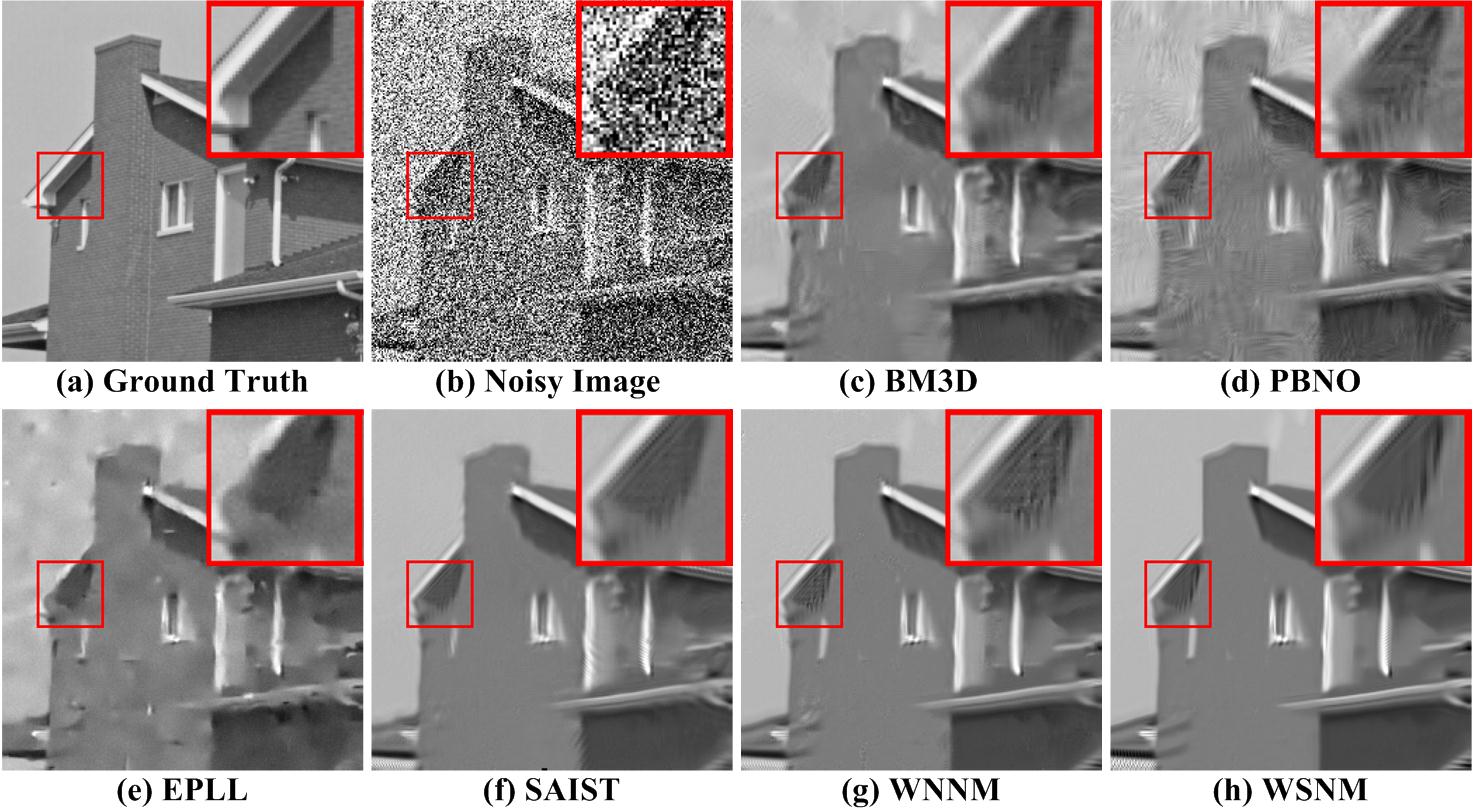}
\caption{Denoising results on image {\it House} at noise level $\sigma_n = 100$. (a) Ground Truth. (b) Noisy Image. (c) BM3D, PSNR = 25.87dB. (d) PBNO, PSNR = 25.42dB. (e) EPLL, PSNR = 25.19dB. (f) SAIST, PSNR = 26.45dB. (g) WNNM, PSNR = 26.67dB. (h) WSNM, PSNR = 26.80dB. The figure is better viewed in zoomed PDF.}
\label{fig:compare-house-sigma100}
\end{figure*}

In this subsection, we compare the proposed WSNM with BM3D, PBNO, SAIST, EPLL, GID and WNNM on 20 widely used test images displayed in Fig. \ref{fig:dataset}. Zero mean additive white Gaussian noises (with variance $\sigma_n = 20, 30, 50, 60, 75, 100$) are added to those test images to generate the noisy observations.

The PSNR performance of seven competing denoising algorithms is reported in Table \ref{comparison-table} (the highest PSNR values are marked in bold). An overall impression observed from Table \ref{comparison-table} is that the proposed WSNM achieves the highest PSNR in almost all cases. When the noise levels are increasing from $20$ to $50$ and to $100$, the improvements of WSNM over WNNM increase from $0.03$dB to $0.1$dB and to $0.12$dB on average, respectively. To sum up, on average our proposed WSNM outperforms all the other competing methods at all noise levels, and the improvement becomes more significant as the noise increases.

In terms of visual quality, as shown in Fig. \ref{fig:compare-girl-sigma50} $\sim$ Fig. \ref{fig:compare-house-sigma100}, our method also outperforms other state-of-the-art denoising algorithms. In the highlighted red window of Fig. \ref{fig:compare-girl-sigma50}, one can see that the proposed WSNM recovers the face structure well, but WNNM generates more artifacts. When we increase the noise level to $100$, it can be seen in the zoom-in window of Fig. \ref{fig:compare-monarch-sigma100} that, the proposed WSNM can well reconstruct wing veins of the butterfly, while many artifacts are produced by other methods. The similar observation is also presented in Fig. \ref{fig:compare-parrot-sigma50} and Fig. \ref{fig:compare-house-sigma100}. In summary, WSNM presents strong denoising capability, producing promising visual quality while keeping higher PSNR indices.

\subsection{Experimental results for WSNM-RPCA}\label{exp-wsnm-rpca}
\subsubsection{\textbf{Synthetic Evaluations}}

\begin{figure*}[htbp]
\setlength{\abovecaptionskip}{0pt}  
\setlength{\belowcaptionskip}{0pt} 
\renewcommand{\figurename}{Figure}
\centering
\includegraphics[width=\textwidth]{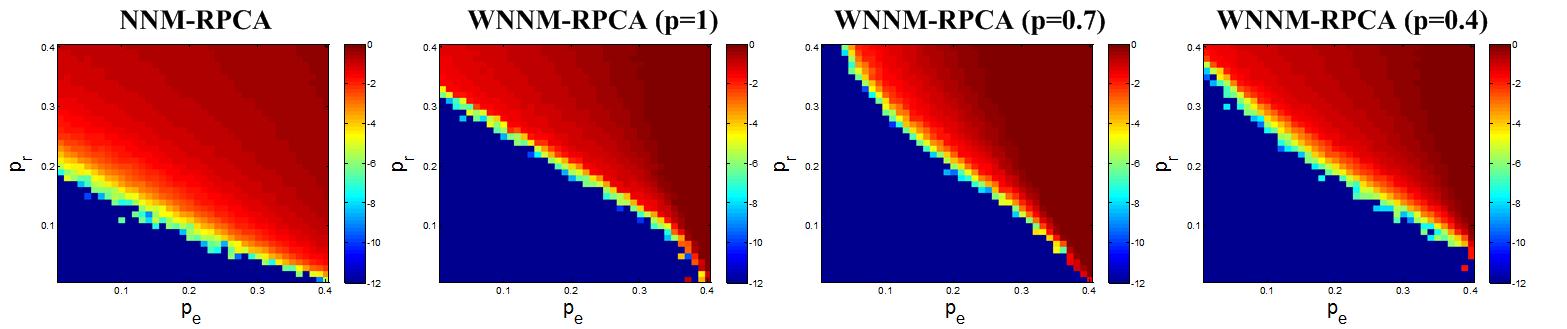}
\caption{The log-scale relative error $\log \frac{\|\hat{X} - X\|_{F}^{2}}{\|X\|_{F}^{2}}$ of NNM-RPCA, WNNM-RPCA and WSNM-RPCA with different ranks and outliers.}
\label{fig:pr-pe-batch-plot}
\end{figure*}

\begin{table*}[!ht]
\footnotesize
\renewcommand{\arraystretch}{1}
\begin{centering}
\begin{threeparttable}[]
\caption{Low rank matrix recovery results by NNM-RPCA, WNNM-RPCA and WSNM-RPCA. $300\times 300$ low rank data with rank from $15$ to $150$; $5\%$ entries are corrupted with sparse noise.}\label{sythetic-exp1}
\tabcolsep=4pt
\begin{minipage}{12cm}
\begin{tabular}{llcccccccccc}
\hline\hline %
& $Rank(\mathbf{X})$ & $15$ & $30$ & $45$ & $60$ & $75$ & $90$ & $105$ & $120$ & $135$ & $150$\\\hline\hline%
\multirow{2}{*}{NNM-RPCA} & $\frac{\|\hat{\mathbf{X}} - \mathbf{X}\|_{F}^{2}}{\|\mathbf{X}\|_{F}^{2}}$
& 2.10E-7 & 3.53E-7 & 3.70E-7 & 6.50E-7 & 3.14E-6 & 0.011 & 0.280 & 0.325 & 0.378 & 0.410\\\cline {2-12}
& $Rank(\hat{\mathbf{X}})$ & 15 & 30 & 45 & 60 & 75 & 98.1 & 160.5 & 159.4 & 155 & 155\\\hline\hline
\multirow{2}{*}{WNNM-RPCA} & $\frac{\|\hat{\mathbf{X}} - \mathbf{X}\|_{F}^{2}}{\|\mathbf{X}\|_{F}^{2}}$
& 3.68E-7 & 6.83E-7 & 1.02E-6 & 1.04E-6 & 1.52E-6 & 0.001 & 0.132 & 0.231 & 0.295 & 0.351\\\cline {2-12}
& $Rank(\hat{\mathbf{X}})$ & 15 & 30 & 45 & 60 & 75 & 90 & 79.5 & 83.1 & 82 & 83.4\\\hline\hline
WSNM-RPCA & $\frac{\|\hat{\mathbf{X}} - \mathbf{X}\|_{F}^{2}}{\|\mathbf{X}\|_{F}^{2}}$
& 3.73E-7 & 7.68E-7 & 2.23E-7 & 7.14E-7 & 1.31E-6 & 1.62E-6 & 1.96E-5 & 0.032 & 0.163 & 0.198\\\cline {2-12}
($p=0.7$) & $Rank(\hat{\mathbf{X}})$ & 15 & 30 & 45 & 60 & 75 & 90 & 105 & 120 & 135 & 150\\\hline\hline
WSNM-RPCA & $\frac{\|\hat{\mathbf{X}} - \mathbf{X}\|_{F}^{2}}{\|\mathbf{X}\|_{F}^{2}}$
& 9.75E-7 & 3.86E-7 & 7.06E-7 & 1.20E-6 & 1.10E-6 & 1.52E-6 & 0.169 & 0.307 & 0.358 & 0.374\\\cline {2-12}
($p=0.4$) & $Rank(\hat{\mathbf{X}})$ & 15 & 30 & 45 & 60 & 75 & 90 & 95.4 & 88 & 87 & 91\\\hline\hline
\end{tabular}
\end{minipage}
\end{threeparttable}
\end{centering}
\end{table*}

\begin{table*}[!ht]
\footnotesize
\renewcommand{\arraystretch}{1}
\begin{centering}
\begin{threeparttable}[]
\caption{Low rank matrix recovery results by NNM-RPCA, WNNM-RPCA and WSNM-RPCA. $300\times 300$ low rank data with rank from $15$ to $150$; $10\%$ entries are corrupted with sparse noise.}\label{sythetic-exp2}
\tabcolsep=4pt
\begin{minipage}{12cm}
\begin{tabular}{llcccccccccc}
\hline\hline %
& $Rank(\mathbf{X})$ & $15$ & $30$ & $45$ & $60$ & $75$ & $90$ & $105$ & $120$ & $135$ & $150$\\\hline\hline%
\multirow{2}{*}{NNM-RPCA} & $\frac{\|\hat{\mathbf{X}} - \mathbf{X}\|_{F}^{2}}{\|\mathbf{X}\|_{F}^{2}}$
& 2.93E-7 & 6.45E-7 & 8.23E-7 & 2.44E-4 & 0.268 & 0.347 & 0.391 & 0.439 & 0.460 & 0.596\\\cline {2-12}
& $Rank(\hat{\mathbf{X}})$ & 15 & 30 & 45 & 60 & 95 & 106 & 155 & 165 & 157 & 157\\\hline\hline
\multirow{2}{*}{WNNM-RPCA} & $\frac{\|\hat{\mathbf{X}} - \mathbf{X}\|_{F}^{2}}{\|\mathbf{X}\|_{F}^{2}}$
& 4.91E-7 & 1.46E-6 & 1.53E-6 & 1.67E-6 & 1.98E-4 & 0.042 & 0.175 & 0.353 & 0.403 & 0.439\\\cline {2-12}
& $Rank(\hat{\mathbf{X}})$ & 15 & 30 & 45 & 60 & 75 & 74 & 74.5 & 79.6 & 80 & 83\\\hline\hline
WSNM-RPCA & $\frac{\|\hat{\mathbf{X}} - \mathbf{X}\|_{F}^{2}}{\|\mathbf{X}\|_{F}^{2}}$
& 5.18E-7 & 9.26E-7 & 1.42E-6 & 1.68E-6 & 1.90E-6 & 0.008 & 0.131 & 0.260 & 0.322 & 0.374\\\cline {2-12}
($p=0.7$) & $Rank(\hat{\mathbf{X}})$ & 15 & 30 & 45 & 60 & 75 & 90 & 105 & 118 & 133 & 149.5\\\hline\hline
WSNM-RPCA & $\frac{\|\hat{\mathbf{X}} - \mathbf{X}\|_{F}^{2}}{\|\mathbf{X}\|_{F}^{2}}$
& 1.03E-6 & 9.27E-7 & 1.30E-6 & 2.04E-6 & 1.86E-6 & 0.070 & 0.220 & 0.314 & 0.421 & 0.437\\\cline {2-12}
($p=0.4$) & $Rank(\hat{\mathbf{X}})$ & 15 & 30 & 45 & 60 & 75 & 90 & 84.4 & 83 & 84 & 86\\\hline\hline
\end{tabular}
\end{minipage}
\end{threeparttable}
\end{centering}
\end{table*}

In this subsection, we conduct extensive synthetic experiments to evaluate the performance of the proposed WSNM-RPCA for low rank data matrix recovery. We generate the ground truth low rank matrix $\mathbf{X}\in \Re^{m \times m}$ by the multiplication of two random matrices generated from a Gaussian distribution $\mathcal{N}(0,1)$, {\it e.g.}, $\mathbf{X} = \mathbf{A}\mathbf{B}^{T}$, where $\mathbf{A}$ and $\mathbf{B}$ are of size $m \times r$. Here $r = m \times \mathcal{P}_r$ indicates the upper bound of $Rank(\mathbf{X})$. Moreover, $m^{2} \times \mathcal{P}_e$ entries in the ground truth matrix $\mathbf{X}$ are corrupted by sparse noise $\mathbf{E}$, whose elements obey a uniform distribution between $[-50, 50]$. In experiments, we set $m=300$, and let $\mathcal{P}_r$ and $\mathcal{P}_e$ vary from $0.01$ to $0.4$ with a step of $0.01$ to test the performance of WSNM-RPCA and its two competitors NNM-RPCA \cite{alm} and WNNM-RPCA (note that WNNM-RPCA is a special case of WSNM-RPCA with $p=1$). For each parameter setting $\{\mathcal{P}_r, \mathcal{P}_e\}$, we generate the synthetic data $10$ times, and the final results are reported by averaging the $10$ runs.

For the NNM-RPCA model, we set $\lambda$, which is an important parameter in optimization, to $1/\sqrt{m}$ to follow the original work \cite{alm}. For WNNM-RPCA and our model, $w_i = C\sqrt{mn}/(\sigma_i(\mathbf{X}) + \varepsilon)$ and we set $C = 10^{1/p}$. The step factor $\rho$ in ALM method for all models is set to $1.2$. Some experimental results are shown in Tables \ref{sythetic-exp1} and \ref{sythetic-exp2}. From them, we can conclude that, when the rank of latent matrix increases or more entries are corrupted in the observed data, the NNM-RPCA always fails to estimate the intrinsic rank of the ground truth matrix. WNNM-RPCA and WSNM-RPCA with $p=0.4$ can get better recovery results than NNM-RPCA but still being sensitive to the rank and noise , while WSNM-RPCA with $p=0.7$ leads to the best recovery of the latent matrix. To facilitate the visual comparison, the log-scale relative error is used to measure the performance, which is defined as:
\begin{equation}\label{log-scale-error}
    \log \frac{\|\hat{\mathbf{X}} - \mathbf{X}\|_{F}^{2}}{\|\mathbf{X}\|_{F}^{2}},
\end{equation}
where $\hat{\mathbf{X}}$ denotes the recovered matrix. The log-scale relative error map of recovered matrix by the three models are presented in Fig. \ref{fig:pr-pe-batch-plot}. From Fig. \ref{fig:pr-pe-batch-plot} one can obviously observe that the success area of WSNM-RPCA (both $p=0.7$ and $p=0.4$) are larger than that of the NNM-RPCA and WNNM-RPCA, which means that WSNM-RPCA is able to recover the low rank matrix with sparse noise in more challenging cases.

\subsubsection{\textbf{Background Subtraction}}

\begin{figure}[!ht]
\setlength{\abovecaptionskip}{0pt}  
\setlength{\belowcaptionskip}{0pt} 
\renewcommand{\figurename}{Figure}
\centering
\includegraphics[width=0.45\textwidth]{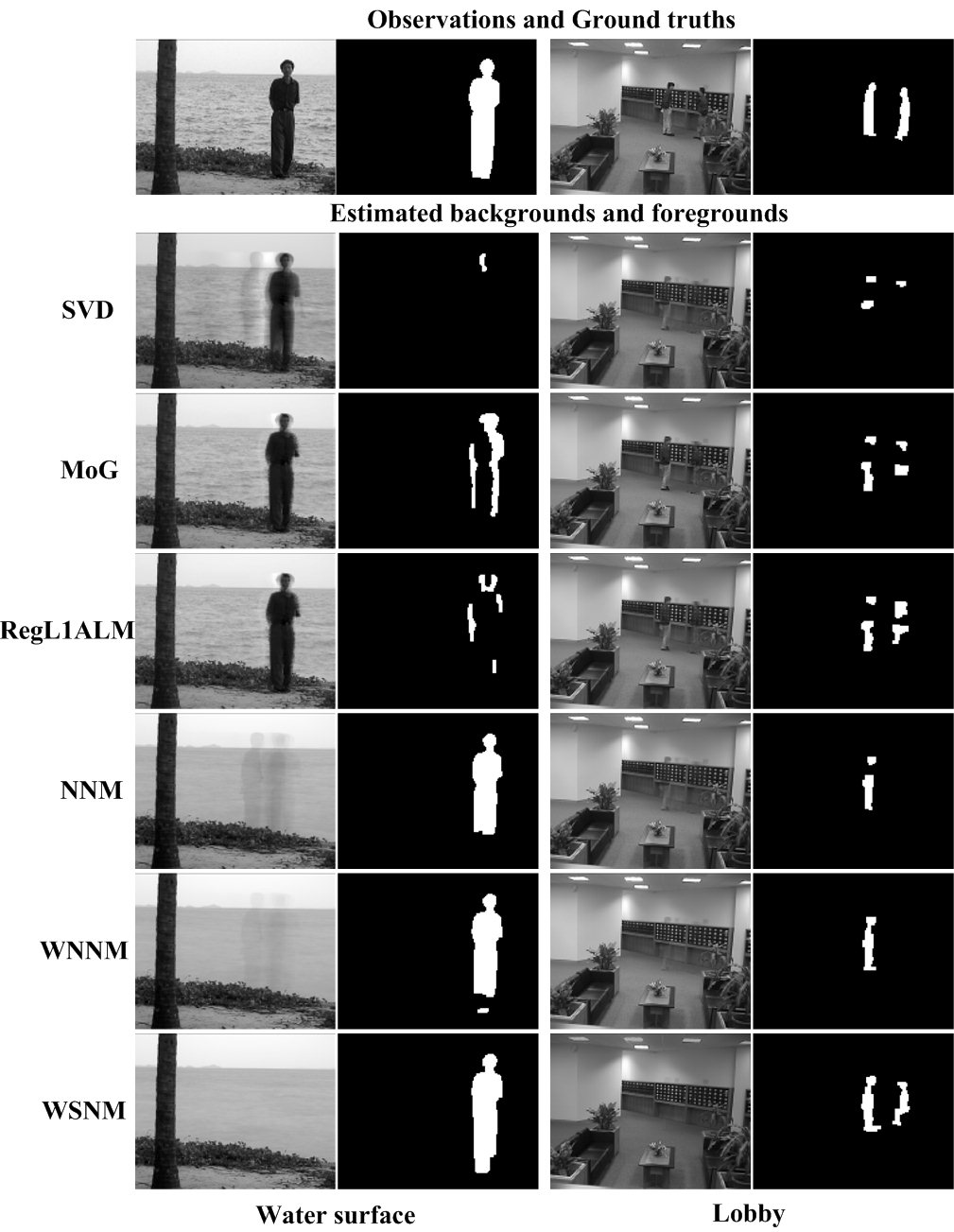}
\caption{The background subtraction results by different methods on {\it Water surface} and {\it Lobby} datasets. The first row is the original frames and their ground truth segmentations.}
\label{fig:background-substraction}
\end{figure}

\begin{table}[htbp]
\footnotesize
\renewcommand{\arraystretch}{1.3}
\begin{centering}
\begin{threeparttable}[]
\caption{Background subtraction results by different methods}\label{bg-comparison}
\tabcolsep=4pt
\begin{minipage}{12cm}
\begin{tabular}{@{}lcccccc}
\toprule[2pt] 
Video Clip         &SVD &MoG\let\thefootnote\relax\footnotetext{\noindent *Note that MoG, NNM, WNNM and WSNM denote MoG-RPCA, \\NNM-RPCA, WNNM-RPCA and WSNM-RPCA, respectively.} &RegL1ALM &NNM &WNNM &WSNM \\ \hline
Watersurface        &0.0995 &0.3194 &0.2231 &0.7703 &0.7884 &\textbf{0.8292} \\
Fountain            &0.2840 &0.6234 &0.4248 &0.5859 &0.6043 &\textbf{0.7329} \\
Lobby              &0.1659 &0.4278 &0.3899 &0.2387 &0.4146 &\textbf{0.6802} \\
Airport            &0.4022 &0.4183 &0.4420 &0.3782 &0.5144 &\textbf{0.6017} \\
Curtain            &0.1615 &0.5675 &0.2983 &0.3191 &0.7634 &\textbf{0.8084} \\
ShoppingMall       &0.3108 &0.4905 &0.5072 &0.4917 &0.5263 &\textbf{0.5788} \\
Campus             &0.3332 &0.3964 &0.3318 &0.3094 &0.5717 &\textbf{0.6093} \\
Bootstrap          &0.3820 &0.4220 &0.3228 &0.3805 &0.4156 &\textbf{0.4576} \\
Escalator          &0.2104 &0.4347 &0.4583 &0.2557 &0.5250 &\textbf{0.5866} \\\hline \bottomrule[1pt]
\end{tabular}
\end{minipage}
\end{threeparttable}
\end{centering}
\end{table}

\begin{figure*}[htbp]
\setlength{\abovecaptionskip}{0pt}  
\setlength{\belowcaptionskip}{0pt} 
\renewcommand{\figurename}{Figure}
\centering
\includegraphics[width=0.8\textwidth]{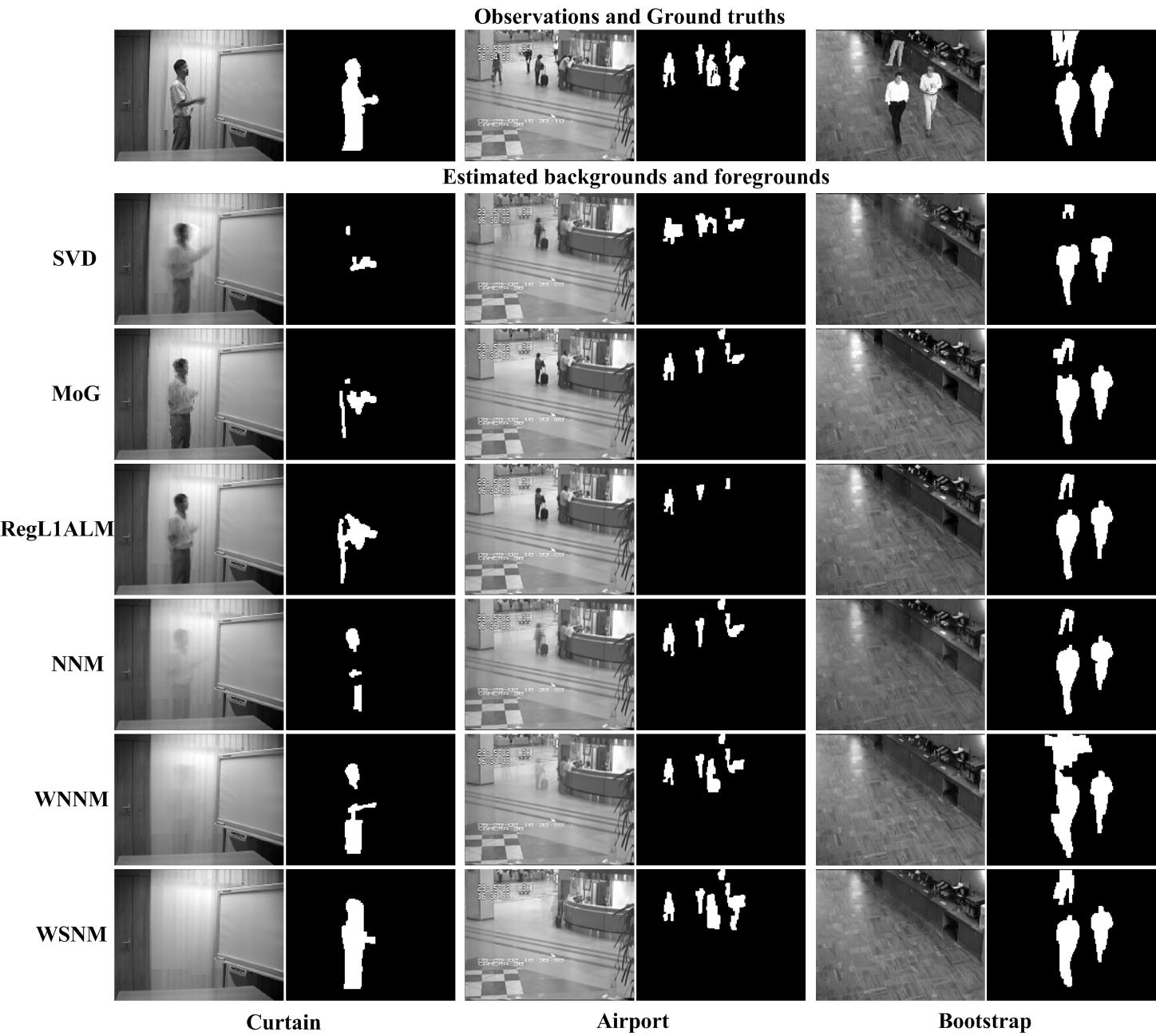}
\caption{The background subtraction results by different methods on {\it Curtain} (left), {\it Airport} (middle) and {\it Bootstrap} (right) datasets. The first row is the original frames and their ground truth segmentations.}
\label{fig:background-substraction-sup}
\end{figure*}

In this subsection, we test the proposed WSNM-RPCA and other competing methods, including NNM-RPCA \cite{alm}, RegL1ALM \cite{LRMAL1}, MoG-RPCA \cite{mogrpca} and WNNM based RPCA, on all the nine video sequences provided by Li {\it et al.} \cite{statistical-model-bg} with {\it all frames} involved. For our WSNM-RPCA, we set the parameter $C$ in Eq.(\ref{wsnm-rpca-weight}) to $2\max(m^2, n^2)$ and power $p = 0.7$. The reason of choosing $p = 0.7$ is based on the experimental results of WSNM-RPCA on synthesis data, as illustrated in Fig. \ref{fig:pr-pe-batch-plot}. It is a tradeoff between enforcing low-rank and separating outliers. To measure the background modeling output quantitatively, we use $S(A, B) = \frac{A\cap B}{A\cup B}$ to calculate the similarity between the estimated foreground regions and the ground truths. To generate the binary foreground map, the MRF model is used to label the absolute value of the estimated sparse error. The quantitative results by different methods are illustrated in Table \ref{bg-comparison}. On all the nine test sequences, the proposed WSNM-RPCA model achieves the best results. Moreover, the visual results of challenging frames in five sequences are shown in Fig. \ref{fig:background-substraction} and Fig. \ref{fig:background-substraction-sup}, which demonstrate that our approach can extract clear background and seperate foreground region with high accuracy. In contrast, the results estimated by other methods exhibit various degrees of ghost shadow in the background, leading to incomplete foreground segmentation.

\section{Conclusions}\label{discussion-and-conclusion}

In this paper, a weighted schatten $p$-norm minimization (WSNM) model was proposed for low rank matrix approximation. WSNM has two major merits: on one hand, it is flexible to fit into practical applications by providing different treatments for different rank components; on the other hand, the schatten $p$-norm promotes the reconstructed low rank matrix to be closer to the latent low rank data matrix. We showed that, when the weights are in non-descending order, the solution of WSNM has global optimum which can be efficiently solved by the generalized iterated shrinkage algorithm. The proposed WSNM was then applied to image denoising and background subtraction to validate its effectiveness. The experimental results demonstrated that WSNM leads to impressive improvements over state-of-the-art methods.

\section{Apendix}\label{proof}

\subsection{Proof of Lemma \ref{solution-lemma}}\label{proof3}
\begin{pf}
  Let the optimal solution of problem (\ref{patch-model1}) have the compact SVD $\mathbf{X} = \mathbf{Q}\mathbf{\Delta} \mathbf{R}^{T}$, and the SVD of matrix $\mathbf{Y}$ be $\mathbf{Y} = \mathbf{U}\mathbf{\Sigma} \mathbf{V}^{T}$, where both $\mathbf{\Delta}$ and $\mathbf{\Sigma}$ are diagonal matrices with the same order (here non-ascending). According to Theorem \ref{von-neumann}, we have
  \begin{equation}
    \begin{aligned}
    &\| \mathbf{X} - \mathbf{Y} \|_{F}^{2} = tr(\mathbf{\Delta}^{T} \mathbf{\Delta}) + tr(\mathbf{\Sigma}^{T} \mathbf{\Sigma}) - 2tr(\mathbf{X}^{T} \mathbf{Y})\\
    &\geq tr(\mathbf{\Delta}^{T} \mathbf{\Delta}) + tr(\mathbf{\Sigma}^{T} \mathbf{\Sigma}) - 2tr(\mathbf{\Delta}^{T} \mathbf{\Sigma}) = \| \mathbf{\Delta} - \mathbf{\Sigma} \|_{F}^{2}.
    \end{aligned}
  \end{equation}
  This implies that
  \begin{equation}\label{}
    \| \mathbf{X} - \mathbf{Y} \|_{F}^{2} + tr(\mathbf{W}\mathbf{\Delta}^{p}) \geq \| \mathbf{\Delta} - \mathbf{\Sigma} \|_{F}^{2} + tr(\mathbf{W}\mathbf{\Delta}^{p}).
  \end{equation}
  Note that the equality holds if and only if $\mathbf{Q} = \mathbf{U}$ and $\mathbf{R} = \mathbf{V}$ according to (\ref{equality-hold}). Therefore, minimizing (\ref{patch-model1}) can be reduced to minimizing the problem in (\ref{equivalence-problem}).
\end{pf}

\subsection{Proof of Lemma \ref{my-lemma}}\label{proof1}
\begin{pf}
  As mentioned in the main paper, for each subproblem, we only need to solve:
    \begin{equation}\label{subproblem-objective}
        f_i(\delta) = \frac{1}{2}(\sigma_i - \delta)^{2} + w_i\delta_i^{p}, 0<p\leq 1, i\in \{1, \ldots, r\}.
    \end{equation}
  According to \cite{gisa}, for any $\sigma_i \in (\tau^{GST}_{p}(w_i), +\infty)$, $f_i(\delta)$ has one unique minimum $S^{GST}_p(\sigma_i;w_i)$, which can be obtained by solving the following equation:
    \begin{equation}\label{implicit-solution}
        S^{GST}_p(\sigma_i;w_i) - \sigma_i + w_i p \bigg(S^{GST}_p(\sigma_i;w_i)\bigg)^{p-1} = 0.
    \end{equation}
  However, solving Eq.(\ref{implicit-solution}) directly is non-trivial, and an iterative algorithm was proposed in \cite{gisa}, which is shown in Algorithm \ref{GST}. The following analysis will be based on this algorithm.

  When $|\sigma| \leq \tau^{GST}_{p}(w_i)$ and $|\sigma| \leq \tau^{GST}_{p}(w_j)$, since $\tau^{GST}_{p}(w)$ is a monotonically increasing function, we have $\tau^{GST}_{p}(w_i) \leq \tau^{GST}_{p}(w_j)$. Then according to Algorithm \ref{GST}, we get $S^{GST}_p(\sigma;w_i) = S^{GST}_p(\sigma;w_j) = 0$. Hence, inequality (\ref{lemma-object}) holds.

  When $|\sigma| > \tau^{GST}_{p}(w_i)$ and $|\sigma| \leq \tau^{GST}_{p}(w_j)$, with Algorithm \ref{GST}, we can achieve $S^{GST}_p(\sigma;w_j) = 0$. Moreover, the object function (\ref{subproblem-objective}) indicates that $S^{GST}_p(\sigma;w_i) \geq 0$ if $\sigma$ is no less than zero, and hence inequality (\ref{lemma-object}) still holds.

  When $|\sigma| > \tau^{GST}_{p}(w_i)$ and $|\sigma| > \tau^{GST}_{p}(w_j)$, we use the mathematical induction method to prove that inequality (\ref{lemma-object}) does hold. Referring to Algorithm \ref{GST}, let $S^{GST}_{p, k}(\sigma;w)$ denote $\delta^{(k)}$ with respect to $w$. When $k=0$, we have $S^{GST}_{p, k}(\sigma;w_i) = S^{GST}_{p, k}(\sigma;w_j) = |\sigma|$, meaning that $S^{GST}_{p,k}(\sigma;w_i) \geq S^{GST}_{p,k}(\sigma;w_j)$ holds. Suppose that inequality $S^{GST}_{p,m}(\sigma;w_i) \geq S^{GST}_{p,m}(\sigma;w_j)$ holds for $k=m$, when $k=m+1$, we have:
  \begin{align}
   S^{GST}_{p,m+1}(\sigma;w_i) &= |\sigma| - w_i p (S^{GST}_{p,m}(\sigma;w_i))^{p-1} \label{equiv_problem3},\\
   S^{GST}_{p,m+1}(\sigma;w_j) &= |\sigma| - w_j p (S^{GST}_{p,m}(\sigma;w_j))^{p-1} \label{equiv_problem4}.
  \end{align}
  Since $S^{GST}_{p,m}(\sigma;w_i) \geq S^{GST}_{p,m}(\sigma;w_j)$ and $0<p\leq 1$, we have:
  \begin{equation}\label{implicit3}
        S^{GST}_{p,m+1}(\sigma;w_i) \geq S^{GST}_{p,m+1}(\sigma;w_j)
  \end{equation}
  So far, we have proven that $S^{GST}_{p,k}(\sigma;w_i) \geq S^{GST}_{p,k}(\sigma;w_j)$ holds for any nonnegative integer $k$. If $k$ reaches $J$, we can also get
  \begin{equation}\label{implicit3}
        S^{GST}_{p}(\sigma;w_i) \geq S^{GST}_{p}(\sigma;w_j),
  \end{equation}
  which means that inequality (\ref{lemma-object}) still holds. This completes the proof of Lemma 2.
\end{pf}

\subsection{Proof of Theorem \ref{my-theorem}}\label{proof4}
\begin{pf}
  Considering the unique minimum $S^{GST}_p(y;w)$ as an implicit function w.r.t. $y$ and $w$, to prove Theorem \ref{my-theorem} we only need to confirm that:
  \begin{equation}\label{GST1}
    S^{GST}_p(y_i;w_i) \geq S^{GST}_p(y_j;w_j), \text{ for } y_i \geq y_j, w_i \leq w_j, i \leq j.
  \end{equation}
  On one hand, for a fixed $w$, the following inequality holds:
  \begin{equation}\label{GST2}
    S^{GST}_p(y_i;w) \geq S^{GST}_p(y_j;w), \text{ for } y_i \geq y_j, i \leq j,
  \end{equation}
  which has been proved in \cite{nikolova}. On the other hand, according to Lemma \ref{my-lemma}, for a fixed $y$, we have
  \begin{equation}\label{GST3}
    S^{GST}_p(y;w_i) \geq S^{GST}_p(y;w_j), \text{ for } w_i \leq w_j, i \leq j.
  \end{equation}
  Therefore, inequalities (\ref{GST2}) and (\ref{GST3}) indicate that (\ref{GST1}) holds. The proof is completed.
\end{pf}

\subsection{Proof of Theorem \ref{theorem-wsnm-rpca}}\label{proof2}
\begin{pf}
  Denote the SVD of matrix $\{\mathbf{Y}+\mu_{k}^{-1}\mathbf{Z}_{k} - \mathbf{E}_{k+1}\}$ in the $k+1$-th iteration as $\mathbf{U}_k \mathbf{\Lambda}_k \mathbf{V}_{k}^{T}$, where $\mathbf{\Lambda}_k$ is the diagonal singular value matrix. By using the GST algorithm for WSNM, we have:
  \begin{equation}\label{}
    \mathbf{X}_{k+1} = \mathbf{U}_k \mathbf{\Delta}_k \mathbf{V}_{k}^{T},
  \end{equation}
  where $\mathbf{\Delta}_k = \{diag(\delta_k^{1}, \delta_k^{2}, \ldots, \delta_k^{n})\}$ is the singular value matrix after generalized soft-thresholding. Thus, based on step 6 in the WSNM-RPCA algorithm, we have:
  \begin{equation}\label{Zk}
  \begin{aligned}
    \|\mathbf{Z}_{k+1}\|_{F}^{2} &= \|\mathbf{Z}_{k} + \mu_{k}(\mathbf{Y} - \mathbf{X}_{k+1} - \mathbf{E}_{k+1})\|_{F}^{2}\\
    &= \mu_{k}^{2}\|\mu_{k}^{-1}\mathbf{Z}_{k} + \mathbf{Y} - \mathbf{X}_{k+1} - \mathbf{E}_{k+1}\|_{F}^{2}\\
    &= \mu_{k}^{2}\|\mathbf{U}_k \mathbf{\Lambda}_k \mathbf{V}_{k}^{T} - \mathbf{U}_k \mathbf{\Delta}_k \mathbf{V}_{k}^{T}\|_{F}^{2}\\
    &= \mu_{k}^{2}\|\mathbf{\Lambda}_k - \mathbf{\Delta}_k\|_{F}^{2}\\
    &\leq \mu_{k}^{2}\|\sum_{i}(\frac{Jw_i}{\mu_k})\|_{F}^{2}\\
    &= \|J\sum_{i} w_i\|_{F}^{2}
  \end{aligned}
  \end{equation}
  where $J$ is the number of iterations in Algorithm GST. So, $\{\mathbf{Z}_k\}$ is bounded.

  From the augmented lagrange function, for a bounded $\mathbf{Z}_k$, $\mathbf{X}_k \rightarrow \infty$ or $\mathbf{E}_k \rightarrow \infty$ will lead to $\mathbf{L}(\mathbf{X},\mathbf{E},\mathbf{Z},\mu) \rightarrow \infty$. Hence, $\mathbf{X}_k$ and $\mathbf{E}_k$ are also bounded. There exist at least one accumulation point for $\{\mathbf{X}_k,\mathbf{E}_k,\mathbf{Z}_k\}$. Specifically, we get
  \begin{equation}\label{}
    \lim_{k\rightarrow\infty} \|\mathbf{Y} - \mathbf{X}_{k+1} - \mathbf{E}_{k+1}\|_{F}^{2} = \lim_{k\rightarrow\infty} \|\mathbf{Z}_{k+1} - \mathbf{Z}_{k}\|_{F}^{2} = 0,
  \end{equation}
  the accumulation point is a feasible solution for the objective function.

  We now prove that the difference between two consecutive iterations will become zero. For $\mathbf{E}$, we have:
  \begin{equation}\label{Ek-part}
  \begin{aligned}
    &\lim_{k\rightarrow\infty} \|\mathbf{E}_{k+1} - \mathbf{E}_k\|_{F}^{2}\\
    &= \lim_{k\rightarrow\infty} \|\mathbf{S}_{\frac{1}{\mu_k}}(\mathbf{Y}+\mu_{k}^{-1}\mathbf{Z}_{k}-\mathbf{X}_{k}) + (\mathbf{Y}+\mu_{k}^{-1}\mathbf{Z}_{k}-\mathbf{X}_{k})\|_{F}^{2}\\
    &\leq \lim_{k\rightarrow\infty} \frac{mn}{\mu_k}=0
  \end{aligned}
  \end{equation}
  where $\mathbf{S}_{\frac{1}{\mu_k}}(\cdot)$ is the soft-thresholding operation with parameter $\frac{1}{\mu_k}$, $m$ and $n$ is the size of the matrix $\mathbf{Y}$. Similarly, for $\mathbf{X}$, we have

  \begin{equation}\label{Xk-part}
  \begin{aligned}
    &\lim_{k\rightarrow\infty} \|\mathbf{X}_{k+1} - \mathbf{X}_k\|_{F}^{2}\\
    &= \lim_{k\rightarrow\infty} \|\mathbf{X}_{k+1} - (\mathbf{Y} + \mu_{k}^{-1}\mathbf{Z}_{k-1} - \mathbf{E}_{k} - \mu_{k}^{-1}\mathbf{L}_{k})\|_{F}^{2}\\
    &\leq \lim_{k\rightarrow\infty} \|\mathbf{X}_{k+1} - (\mathbf{Y} - \mathbf{E}_{k+1})\|_{F}^{2} + \|\mu_{k}^{-1}\mathbf{Z}_{k-1} - \mu_{k}^{-1}\mathbf{Z}_{k}\|_{F}^{2}\\
    &=\lim_{k\rightarrow\infty} \|\mathbf{U}_{k} (\mathbf{\Delta}_{k} - \mathbf{\Sigma}_{k}) \mathbf{V}_{k}^{T} + \mu_{k}^{-1}\mathbf{Z}_{k}\|_{F}^{2} + \mu_{k}^{-2}\|\mathbf{Z}_{k-1} - \mathbf{Z}_{k}\|_{F}^{2}\\
    &\leq \|\sum_{i}(\frac{Jw_i}{\mu_k})\|_{F}^{2} + \mu_{k}^{-2}(\|\mathbf{Z}_{k}\|_{F}^{2} + \|\mathbf{Z}_{k-1} - \mathbf{Z}_{k}\|_{F}^{2}) = 0.
  \end{aligned}
  \end{equation}
  The proof is completed.
\end{pf}

{\small
\bibliographystyle{unsrt}
\bibliography{egbib}
}
\end{document}